\pdfoutput=1

\documentclass[letterpaper, 10 pt, conference]{ieeeconf}  % Comment this line 
\IEEEoverridecommandlockouts                              % This command is only needed if 
\usepackage{graphicx}
\usepackage{caption}
\usepackage{bm}  % Bold
\usepackage{booktabs}
\usepackage{amsmath} % for equation 
\usepackage{array}  % adjust the column of the table
\usepackage{multirow}
\usepackage{stfloats}
\usepackage{algorithmicx,algorithm}
\usepackage{algpseudocode}
\usepackage{float}
\usepackage{subcaption}
\newsavebox{\imagebox}
\usepackage{amsfonts}
\usepackage{tikz}
\usepackage{stackengine}
\usepackage{makecell}

\newcolumntype{L}[1]{>{\raggedright\let\newline\\\arraybackslash\hspace{0pt}}m{#1}}
\newcolumntype{C}[1]{>{\centering\let\newline\\\arraybackslash\hspace{0pt}}m{#1}}
\newcolumntype{R}[1]{>{\raggedleft\let\newline\\\arraybackslash\hspace{0pt}}m{#1}}

\newcolumntype{M}[1]{>{\centering\arraybackslash}m{#1}}

\definecolor{elbow1_color}{rgb}{0.4392,0.6196,0.9765}
\definecolor{elbow2_color}{rgb}{0.9882, 0.4863, 0}
\definecolor{wrist1_color}{RGB}{135, 206, 250}
\definecolor{wrist2_color}{RGB}{127, 255, 212}
\definecolor{ee_color}{RGB}{0, 191, 255}
\definecolor{SLSQP_Color}{RGB}{200, 212, 240}

\newcommand{\tikzcircle}[2][red,fill=red]{\tikz[baseline=-0.7ex]\draw[#1,radius=#2] (0,0) circle ;}%

\title{\LARGE \bf A Combined Inverse Kinematics Algorithm Using FABRIK with Optimization}

\author{Zichun Xu$^{1}$, Yuntao Li$^{1}$, Xiaohang Yang$^{1}$, Zhiyuan Zhao$^{1}$, Jingdong Zhao$^{1}$, and Hong Liu$^{1}$% <-this % stops a space
\thanks{This work has been supported by the National Natural Science Foundation of China [Project Number: 92148203], the State Key Laboratory of Robotics and System (HIT) [Project Number: SKLRS202201A01], and the Key Lab. of Science and Technology on Space Flight Dynamics [Project Number: XTB6142210210303]. (\emph{Corresponding author: Jingdong Zhao.})}% <-this % stops a space
\thanks{$^{1}$All authors are with the State Key Laboratory of Robotics and System, Harbin Institute of Technology, Harbin 150001, Heilongjiang Province, China.}%
}



\begin{document}
\maketitle
\thispagestyle{empty}
\pagestyle{empty}

%\listoffigures
\captionsetup[figure]{labelformat={default},labelsep=period,name={Fig.}} % labelfont={bf}

%%%%%%%%%%%%%%%%%%%%%%%%%%%%%%%%%%%%%%%%%%%%%%%%%%%%%%%%%%%%%%%%%%%%%%%%%%%%%%%%
\begin{abstract}
	Forward and backward reaching inverse kinematics (FABRIK) is a heuristic inverse kinematics solver that is gradually applied to manipulators with the advantages of fast convergence and generating more realistic configurations. However, under the high error constraint, FABRIK exhibits unstable convergence behavior, which is unsatisfactory for the real-time motion planning of manipulators. In this paper, a novel inverse kinematics algorithm that combines FABRIK and the sequential quadratic programming (SQP) algorithm is presented, in which the joint angles deduced by FABRIK will be taken as the initial seed of the SQP algorithm to avoid getting stuck in local minima. The combined algorithm is evaluated with experiments, in which our algorithm can achieve higher success rates and faster solution times than FABRIK under the high error constraint. Furthermore, the combined algorithm can generate continuous trajectories for the UR5 and KUKA LBR IIWA 14 R820 manipulators in path tracking with no pose error and permitted position error of the end-effector.
\end{abstract}


%%%%%%%%%%%%%%%%%%%%%%%%%%%%%%%%%%%%%%%%%%%%%%%%%%%%%%%%%%%%%%%%%%%%%%%%%%%%%%%%
\section{Introduction}
	Inverse kinematics (IK), which is typically applied to robotics and computer graphics, is a nonlinear mapping from the end-effector (EE) to each degree of freedom (DOF). Some IK algorithms can be effective in both areas, such as the Jacobian transpose \cite{chiacchio1989closed}, Jacobian pseudo-inverse \cite{wang2010inverse}, and Cyclic Coordinate Descent (CCD) \cite{wang1991combined}. Recently, an IK algorithm called Forward And Backward Reaching Inverse Kinematics (FABRIK) \cite{aristidou2011fabrik} has been developed to solve the IK problems of the articulated body and kinematic chain with multiple EEs. Then, FABRIK is extended with model constraints and more types of joints \cite{aristidou2016extending}. More cases are considered to improve the flexibility of FABRIK. Even when the target is unreachable, a suitable solution can also be given with one single iteration. Many studies have proved FABRIK's excellent capabilities in low computation cost, generating smooth motions, redundancy resolution, and dealing with singularities, implying that it has tremendous potential for application to robots. 
	
\begin{figure}[t]
	\centering
	\includegraphics[scale=1.]{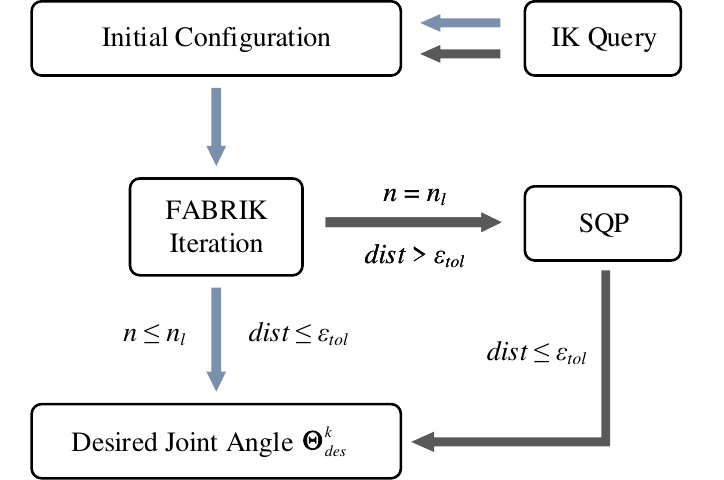}
	\caption{Scheme of the combined algorithm. FABRIK and the SQP algorithm are integrated and switched based on the detection condition. Given the IK query and the initial configuration of the kinematic chain, if FABRIK can converge within limited iterations $ \big($i.e., $n \le n_l$ and  $dist \le \varepsilon_{tol} \big)$, the desired joint angle vectors $ \bm{\Theta}_{des}^k $ will be obtained through analytical derivation. Alternatively, the optimization method will be activated.}
	\label{all_procedure}
\end{figure}

	However, solving the IK problems for robots is generally different from that in animation. For robot EE, a higher Cartesian error (position/pose error) constraint is required, and joint angles must be derived in addition to joint Cartesian positions, which require the IK algorithm to maintain a high success rate and computational efficiency. Theoretically, the practical Cartesian error requirement is $ 10^{-6} $, as the setting of quantitative tests in \cite{beeson2015trac}. However, under this Cartesian error constraint, the stability of FABRIK fluctuates greatly, i.e., FABRIK consumes a large amount of time for convergence in some cases. For these cases, FABRIK approaches the target with high efficiency at the beginning of the iterations and corrects the joint positions at a relatively slow rate in the following iterations. To address this problem and extend FABRIK to manipulators, a novel IK algorithm that combines FABRIK and an optimization algorithm is presented to obtain excellent convergence stability and computation efficiency when applied to manipulators. The scheme of the combined algorithm is summarized in Fig.~\ref{all_procedure}.
\subsection{Related Work}
Due to an abundance of relevant studies about IK and the focus of this paper, a brief review of FABRIK and optimization algorithms is presented in this subsection.

Heuristic algorithms are the popular IK algorithms with low computational cost during iteration, where CCD and FABRIK are the most representative ones. From EE to base, CCD iteratively changes the angle of each joint to approach the target. However, CCD may cause abrupt movements with oscillations and unnatural configurations. The natural-CCD \cite{martin2018natural} fixes the limitation of CCD and provides real-time IK calculations for hyper-redundant manipulators. FABRIK \cite{aristidou2011fabrik, aristidou2016extending}, which is a recent algorithm, divides the iteration process into the forward and backward phases, where joint Cartesian positions are corrected repeatedly according to the target and base, respectively. In contrast to CCD, FABRIK can generate better whole-body movements and natural configurations, especially for humanoid models and serial kinematic chains. Ananthanarayanan and Ordóñez \cite{ana2015real} use FABRIK to deduce the feasible configuration of the manipulator with $ (2n+1) $ DOFs, which is utilized to optimize elbow positions to avoid obstacles. Using FABRIK, Tao and Yang \cite{tao2017collision} propose a collision avoidance motion planning method for virtual arms. The directions of links are adjusted during the iteration phase to avoid obstacles and eventually deduce a collision-free configuration in the plane \cite{tao2021extending}. Dong et al. \cite{gangqi2021modified} impose the joint velocity limits in the backward phase and adjust the joint positions after iterations based on the momentum conservation law when applied to the space manipulator with a floating base. Kolpashchikov et al. \cite{kolpashchikov2018fabrik} test the FABRIK-based algorithm for solving the IK of multi-section continuum robots with the constraint of a $ 100\ \mu m $ position error and a ${0.1^ \circ}$ pose error. Although many FABRIK-based IK algorithms \cite{santos2020m, xie2019obstacle} have been developed for application to different types of manipulators, none of them have been quantitatively tested under high Cartesian error constraints. Since the performance of these algorithms is essentially determined by FABRIK, the convergence of FABRIK under high error requirements needs to be analyzed and further optimized.

For the IK of robots, which can be formulated as nonlinear optimization problems, the optimization method is a suitable option and can better handle several constraints, such as the genetic algorithm \cite{nearchou1998solving}, particle swarm optimization \cite{ram2019inverse}, and Broyden–Fletcher–Goldfarb–Shanno (BFGS) \cite{fletcher2013practical}. Starke et al. \cite{starke2018memetic} combine these three algorithms and propose a novel memetic evolutionary algorithm that performs well on different robots integrated with additional task constraints. Mari{\'c} et al. \cite{maric2021riemannian} construct the kinematic model with additional constraints based on distance geometry and deduce the IK solutions for different manipulators using Riemannian optimization. Instead of solving IK problems for the trajectory points, Shirafuji and Ota \cite{shirafuji2019kinematic} use the derivate-free optimization method to minimize the tracking error of the target and desired trajectories. Another well-known optimization method is the sequential quadratic programming (SQP) algorithm, which can use BFGS to iteratively search for the solution \cite{kumar2010optimization}. For instance, Beeson and Ames \cite{beeson2015trac} construct the TRAC-IK solver, which combines a series of SQP variants and the Orocos Kinematics and Dynamics Library, to achieve considerable success rates and low computation costs on different robot platforms. Lyu et al. \cite{lyu2017time} use BFGS to update the Hessian matrix and generate the time-optimal trajectory with energy optimization. However, SQP is susceptible to initial value and biases the search around it. Inappropriate initial values can make SQP get stuck in a local minimum. Xie et al. \cite{xie2022speedup} speed up SQP by sampling a large amount of data and selecting the one that is closest to the desired pose as the initial seed. Although it is an inefficient method to determine the initial value of the SQP algorithm, it indicates that the initial value has a significant impact on convergence.

\subsection{Contribution and Organization}
	Based on the previous work, it can be concluded that SQP algorithms may converge faster than heuristic algorithms in the presence of gradient information, but will be influenced by the initial value. Thus, this paper presents an IK algorithm that exploits the advantages of FABRIK and the SQP algorithm, which is implemented by the SLSQP algorithm of the NLopt library. As shown in Fig.~\ref{all_procedure}, this algorithm is mainly divided into two procedures, wherein FABRIK is performed first within a limited number of iterations. If FABRIK fails to converge with the given Cartesian error constraint, SLSQP is then executed, and the approximate solution generated by FABRIK is used as the initial value of SLSQP. The main contributions of this paper are as follows:

\begin{enumerate}
	\item The convergence property of FABRIK is discussed in detail and summarized. 
	
	\item A novel IK algorithm that combines FABRIK and the SQP algorithm is presented to compensate for the slow convergence rate of FABRIK in some two-dimensional (2-D) and three-dimensional (3-D) cases. 
	
	\item In contrast to FABRIK, quantitative tests are performed on the UR5 and KUKA LBR IIWA 14 R820 manipulators to demonstrate the ability of the combined algorithm to provide high solve rates and low computation costs with the $ 10^{-6} $ Cartesian error constraint.
	
	\item Experiments are conducted to substantiate the effectiveness of the combined algorithm in path tracking and ensure the accuracy of EE. 
\end{enumerate}

The remainder of this paper is organized as follows: Sec.~\ref{sec2} first reviews FABRIK and analyzes its unstable convergence property in some 2-D and 3-D scenarios. Sec.~\ref{sec3} introduces the combined algorithm and the procedures for applying it to the UR5 and KUKA manipulators. Sec.~\ref{sec4} compares FABRIK and the combined algorithm through convergence comparison and quantitative tests, and then evaluates this algorithm with path tracking tasks. Finally, Sec.~\ref{sec6} concludes this paper with some future suggestions.

\section{FABRIK convergence analysis}
\label{sec2}
\begin{figure}[t]
	\centering
	\captionsetup[subfigure] {skip=0pt,slc=off,margin={60pt, 0pt},labelfont=normalfont}
	%	\subcaptionbox{\label{link-1}}[4cm][c]{\includegraphics[scale=1]{link1.pdf}}
	%	\subcaptionbox{\label{link-2}}[4cm][c]{\includegraphics[scale=1]{link2.pdf}}
	\subcaptionbox{\label{link-1}}[4cm][c]{\includegraphics[scale=0.19]{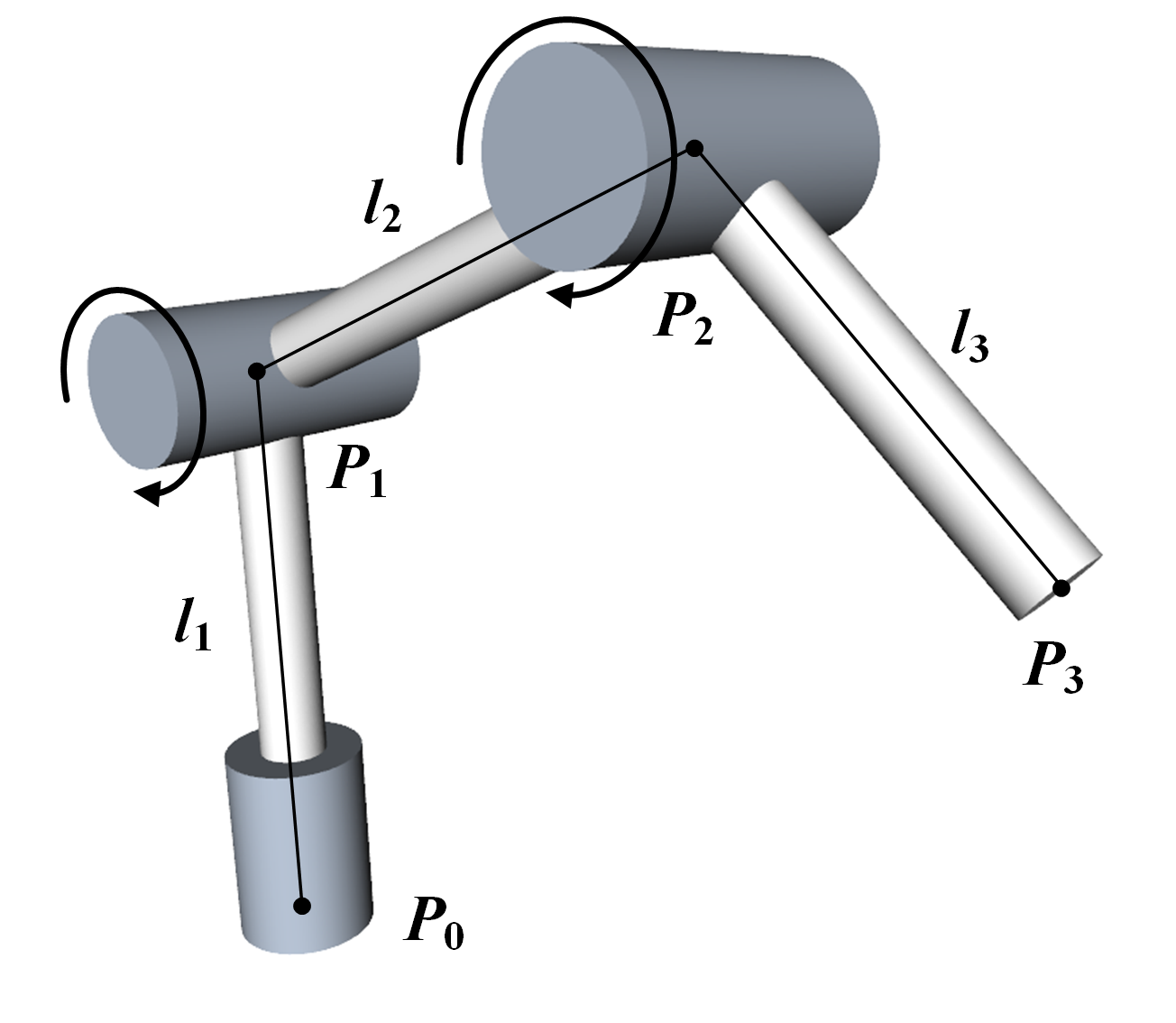}}
	\subcaptionbox{\label{link-2}}[4cm][c]{\includegraphics[scale=0.19]{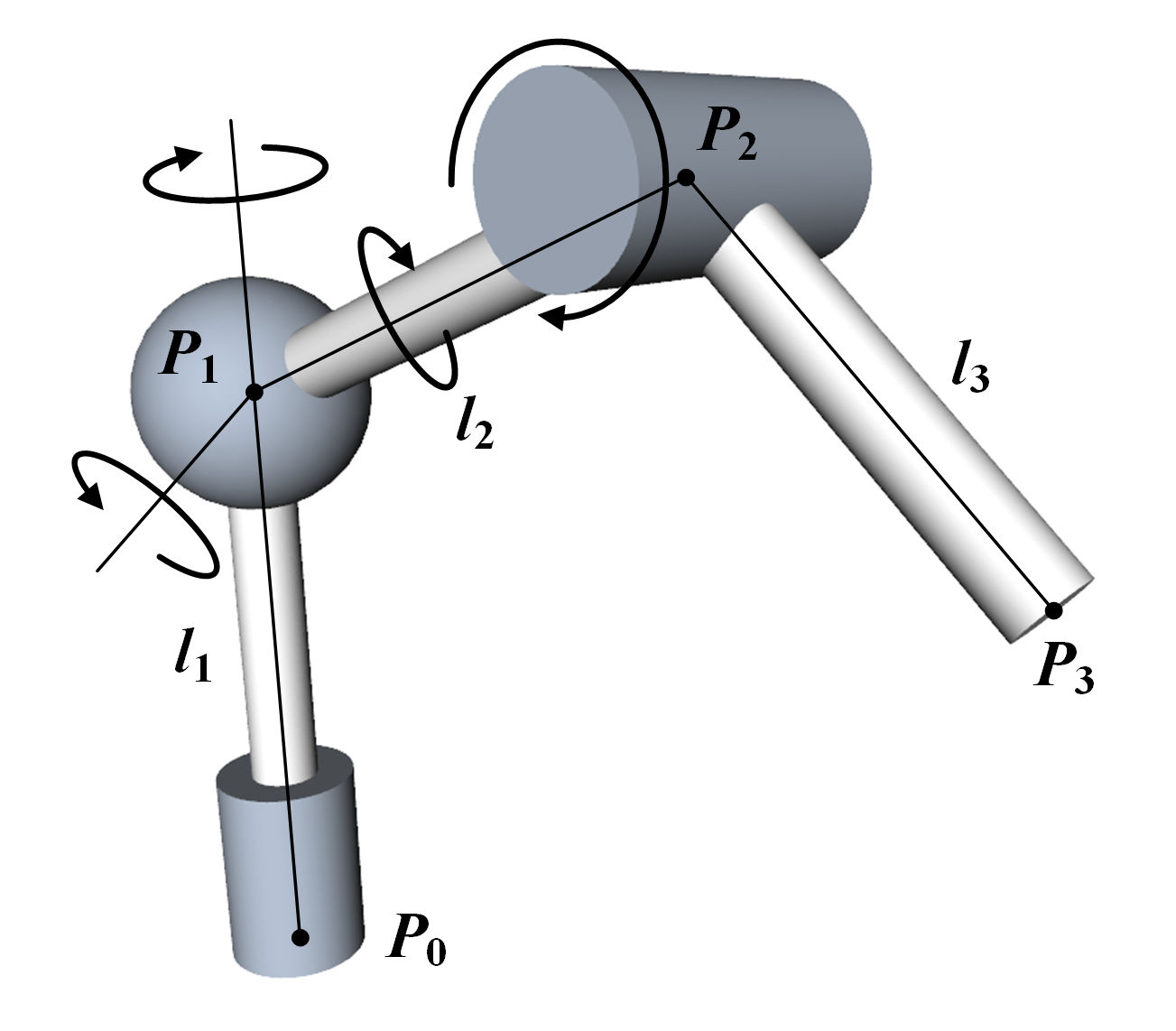}}
	\caption{Schematics of the (\subref{link-1}) planar 2-DOF and (\subref{link-2}) spherical-revolute 4-DOF manipulators.}
	\label{links}
\end{figure}

In this section, the iteration processes of FABRIK in 2-D and 3-D scenarios are first reviewed and described by mathematical expressions. In addition, the convergence property of FABRIK is discussed in detail and summarized based on the two types of manipulators in Fig.~\ref{links}. All the calculations and experiments in this paper are implemented in C++ on a computer with an Intel Core i7-12700K CPU and 64 GB RAM. Tab.~\ref{notation} summarizes some important mathematical notations used in this paper.

\begin{figure*}[t!]
	\captionsetup[subfigure] {skip=-105pt,slc=off,margin={0pt, 0pt},labelfont=normalfont} 
	\centering
	%		\subcaptionbox{\label{2d-1}}[4.2cm][c]{\includegraphics[width=4.2cm]{D:/Dr File/ICRA2022/EI_2022/UR5_init.eps}}
	\subcaptionbox{\label{2d-1}}[4.5cm][c]{\includegraphics[width=4.5cm]{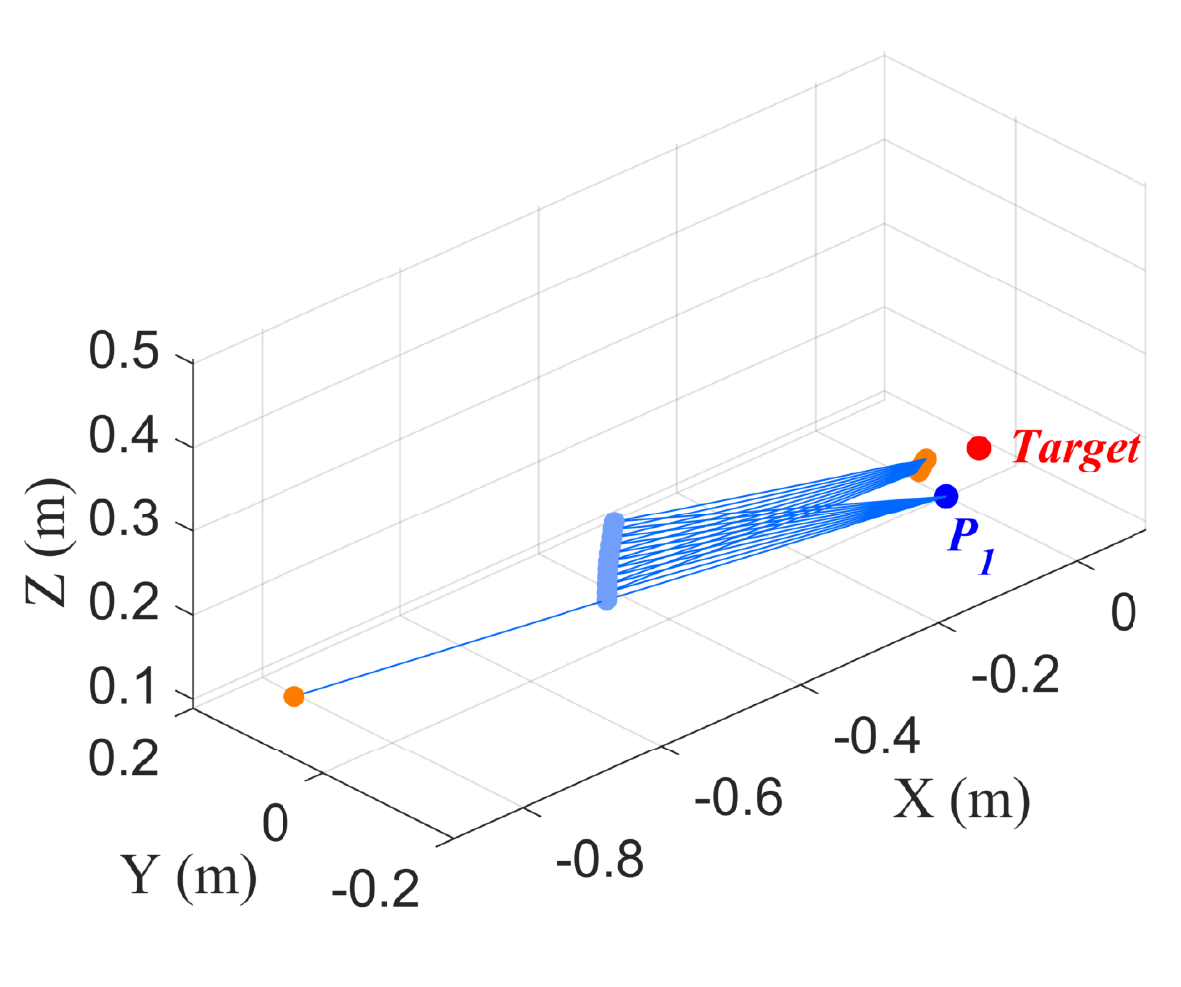}}
	\hspace{0.2cm}
	\subcaptionbox{\label{2d-2}}[4.5cm][c]{\includegraphics[width=4.5cm]{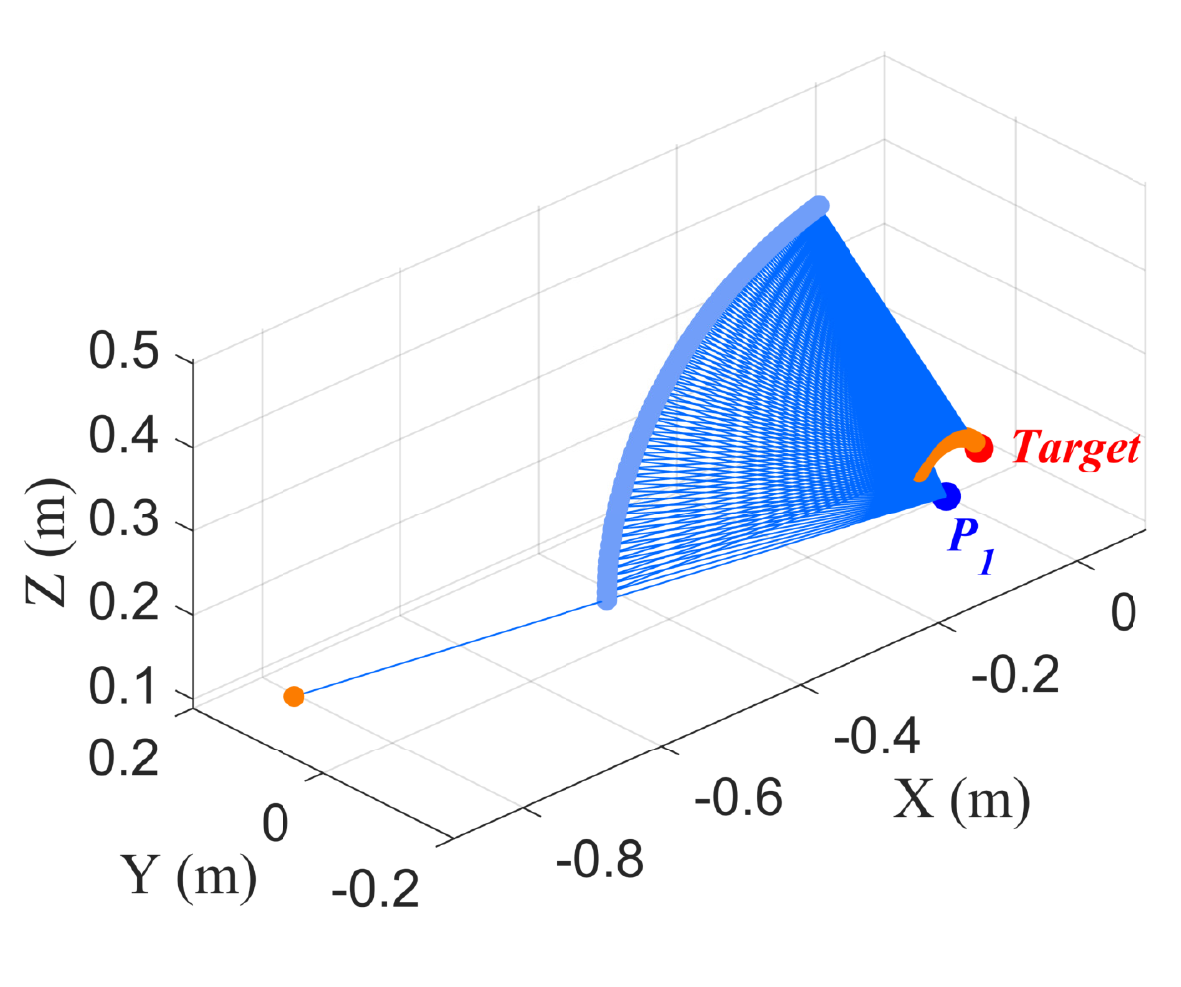}}
	\hspace{0.2cm}
	\subcaptionbox{\label{2d-3}}[4.5cm][c]{\includegraphics[width=4.5cm]{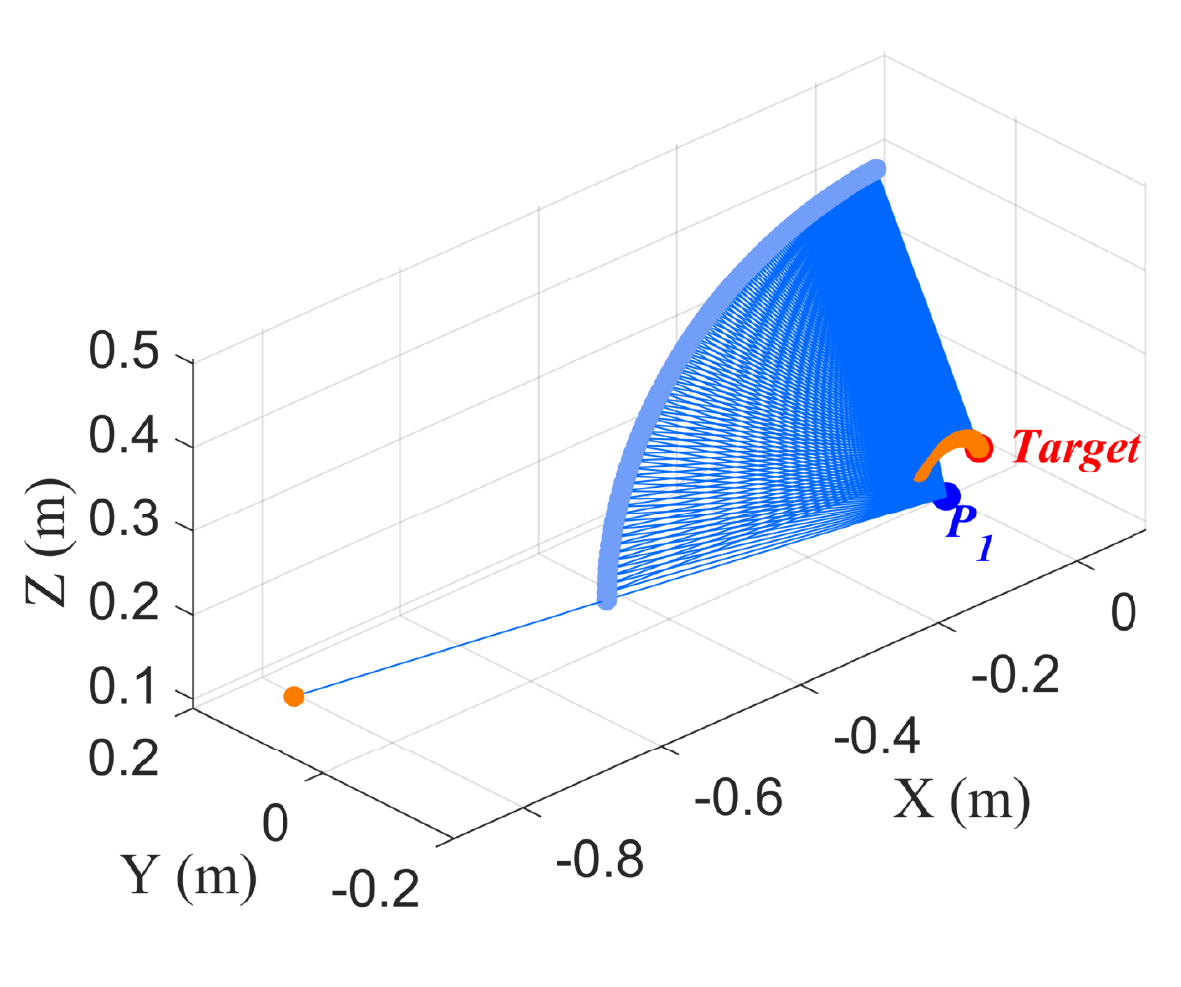}}
	
	\subcaptionbox{\label{2d-4}}[4.5cm][c]{\includegraphics[width=4.5cm]{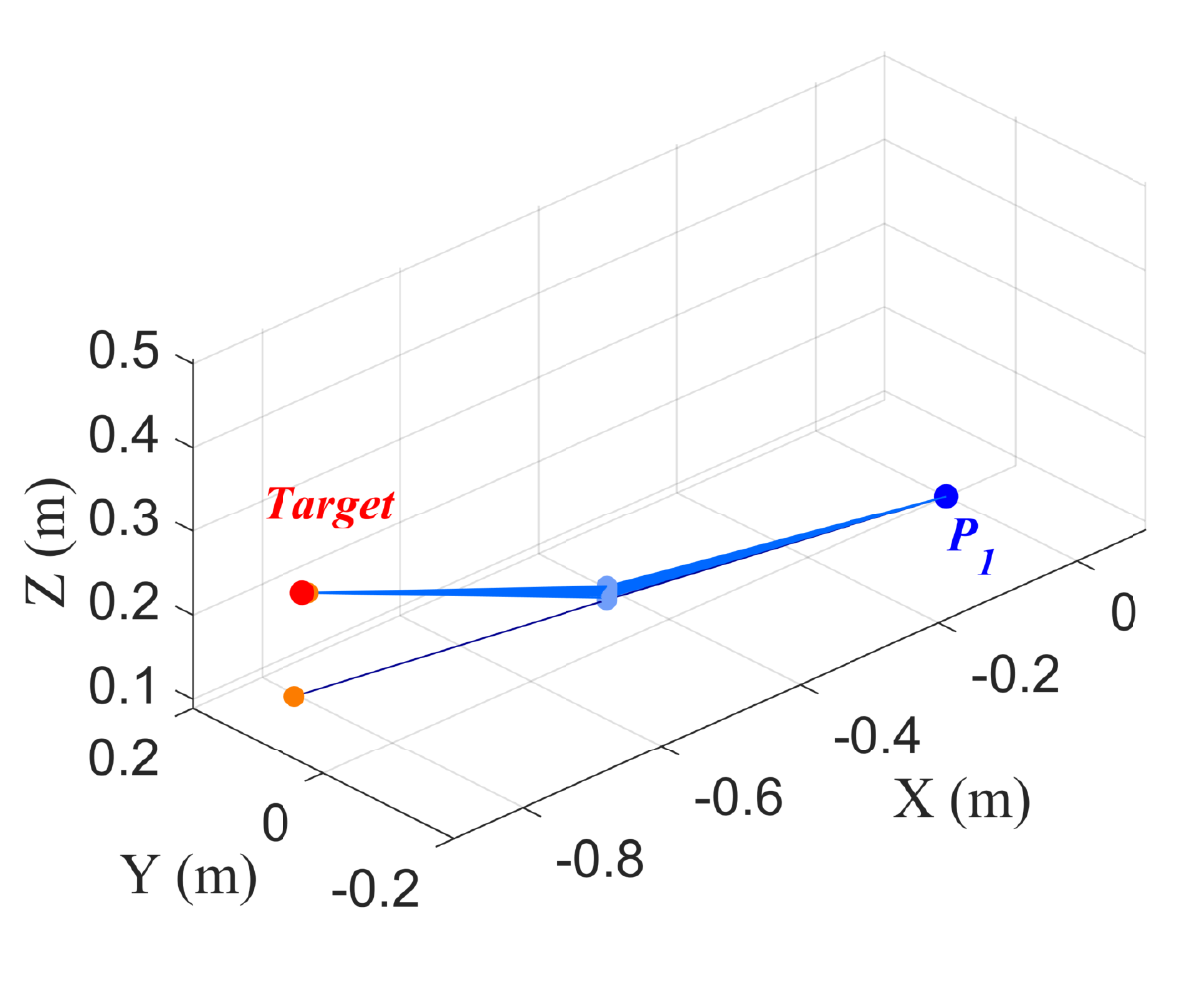}}
	\hspace{0.2cm}
	\subcaptionbox{\label{2d-5}}[4.5cm][c]{\includegraphics[width=4.5cm]{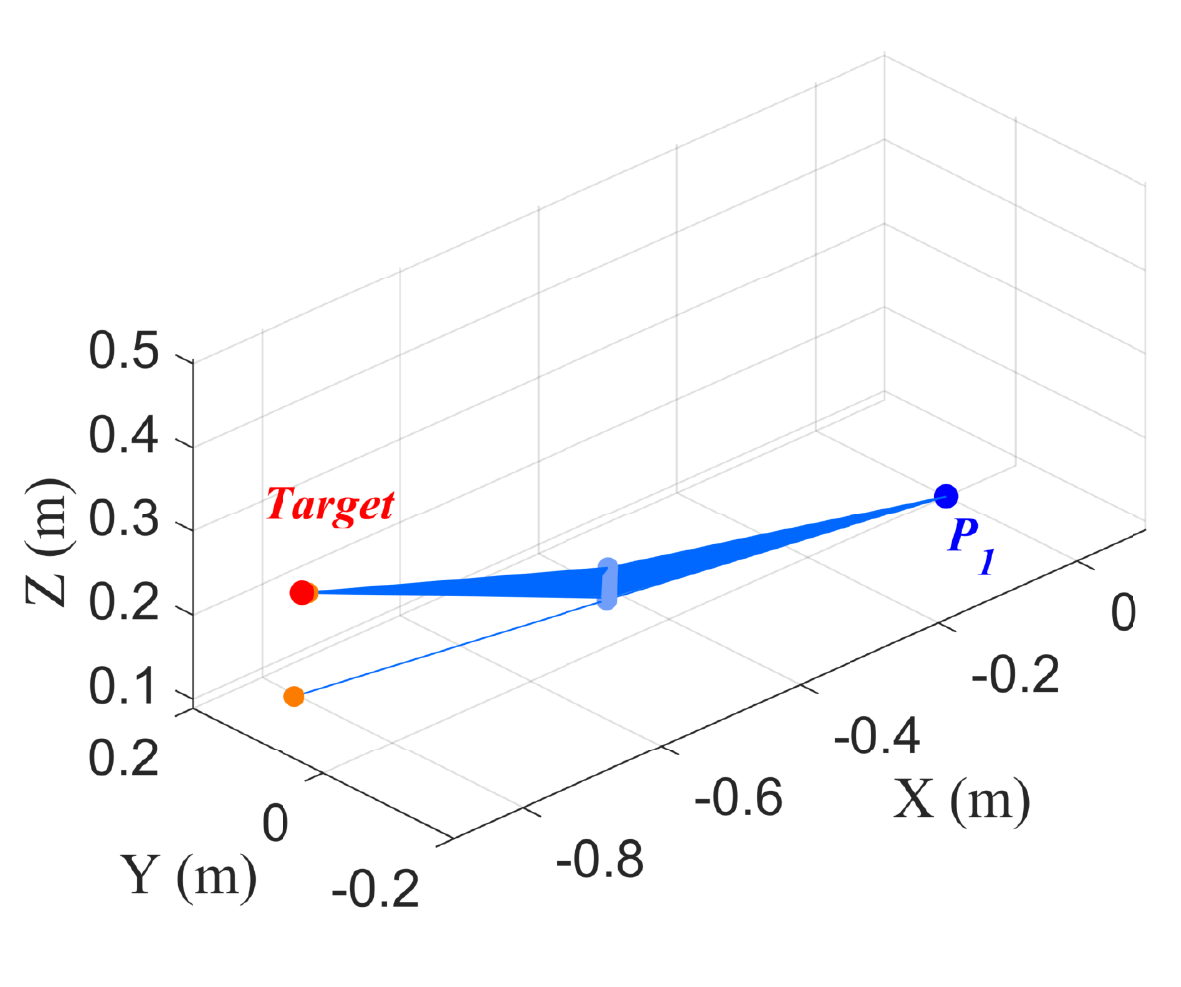}}
	\hspace{0.2cm}
	\subcaptionbox{\label{2d-6}}[4.5cm][c]{\includegraphics[width=4.5cm]{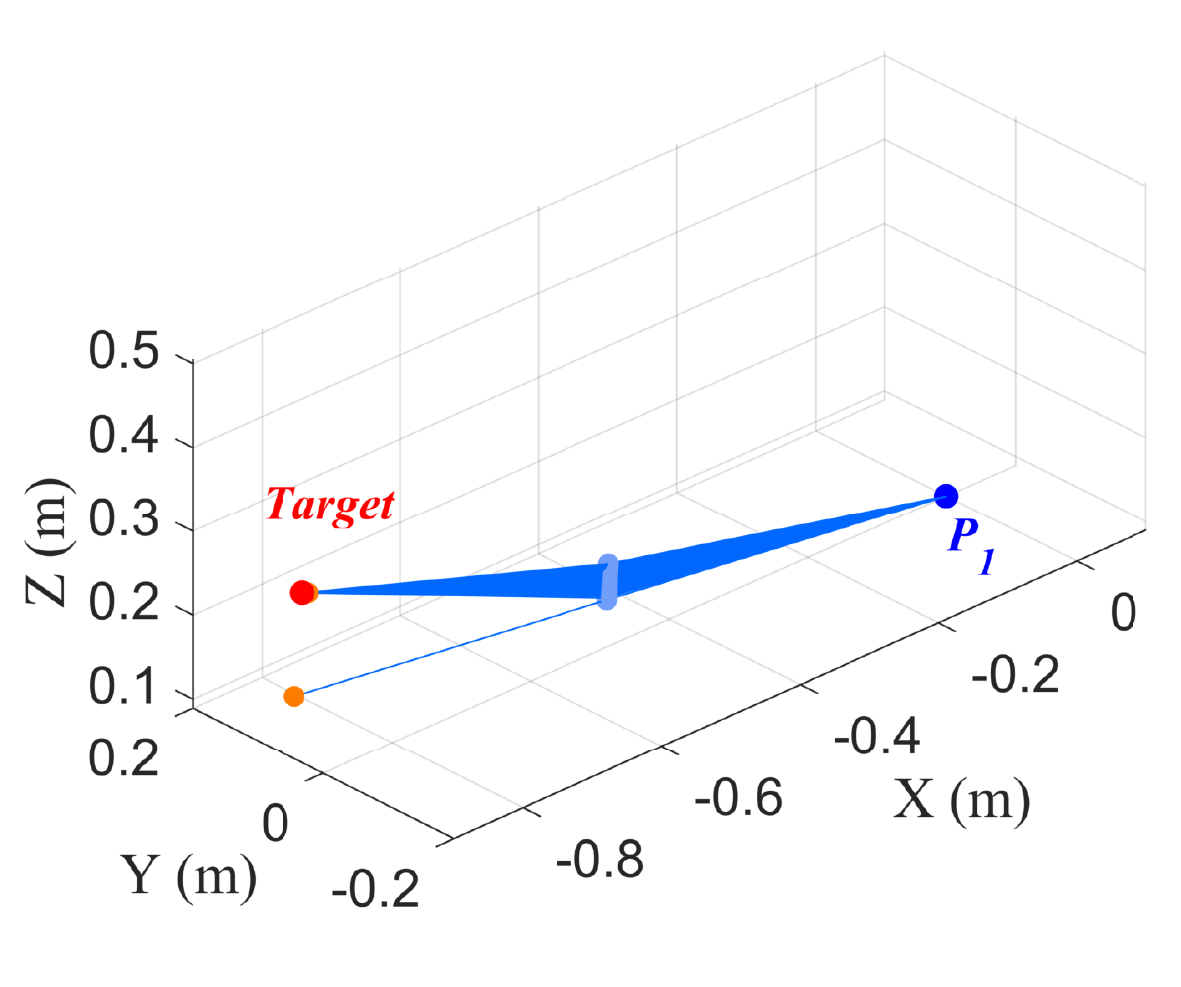}}
	
	%	\vskip -pt
	%		\subcaptionbox{\label{2d-5}}[4.2cm][c]{\includegraphics[width=4.2cm]{D:/Dr File/ICRA2022/EI_2022/KUKA_init.eps}}
	\subcaptionbox{\label{2d-7}}[4.5cm][c]{\includegraphics[width=4.5cm]{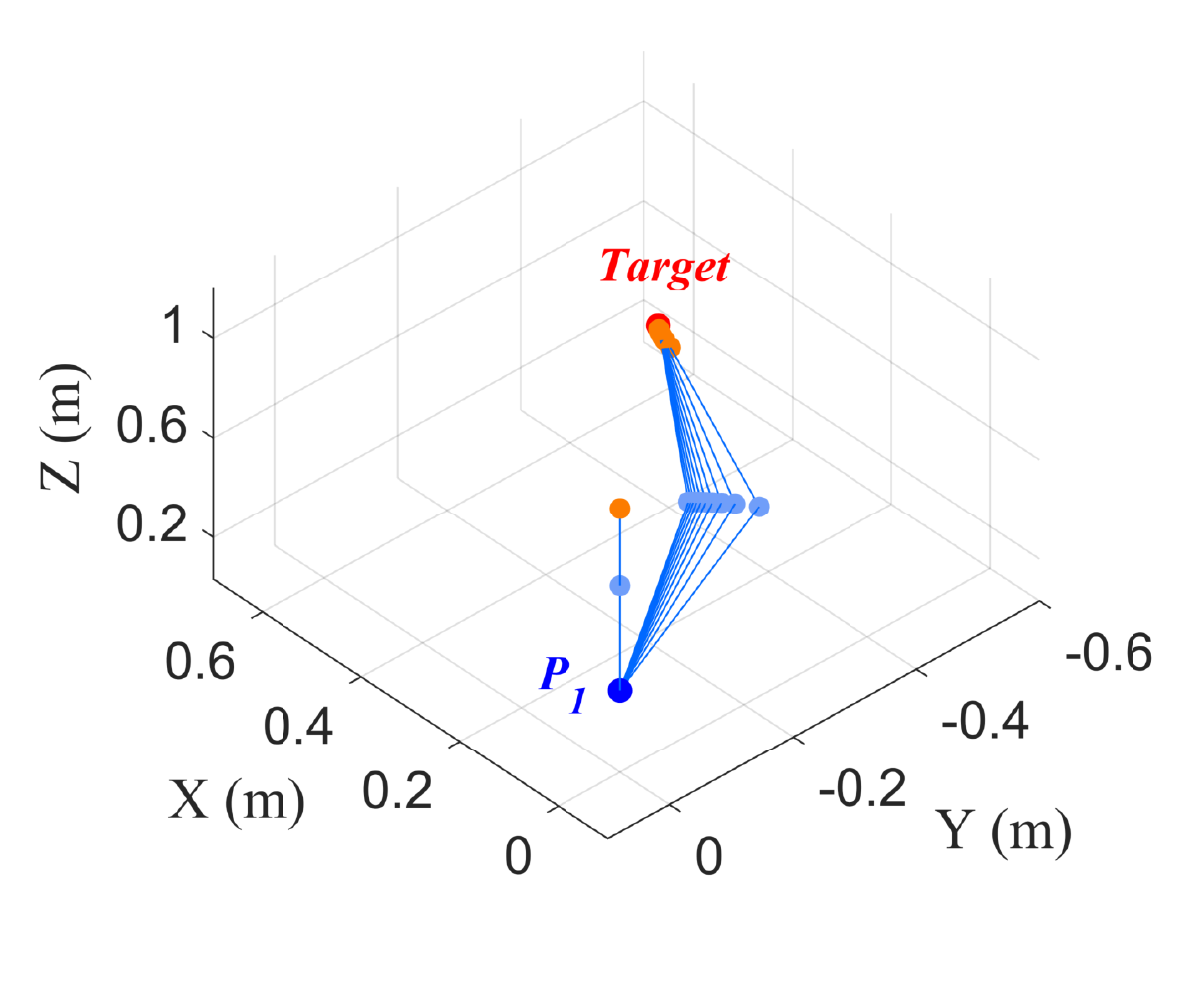}}
	\hspace{0.3cm}
	\subcaptionbox{\label{2d-8}}[4.5cm][c]{\includegraphics[width=4.5cm]{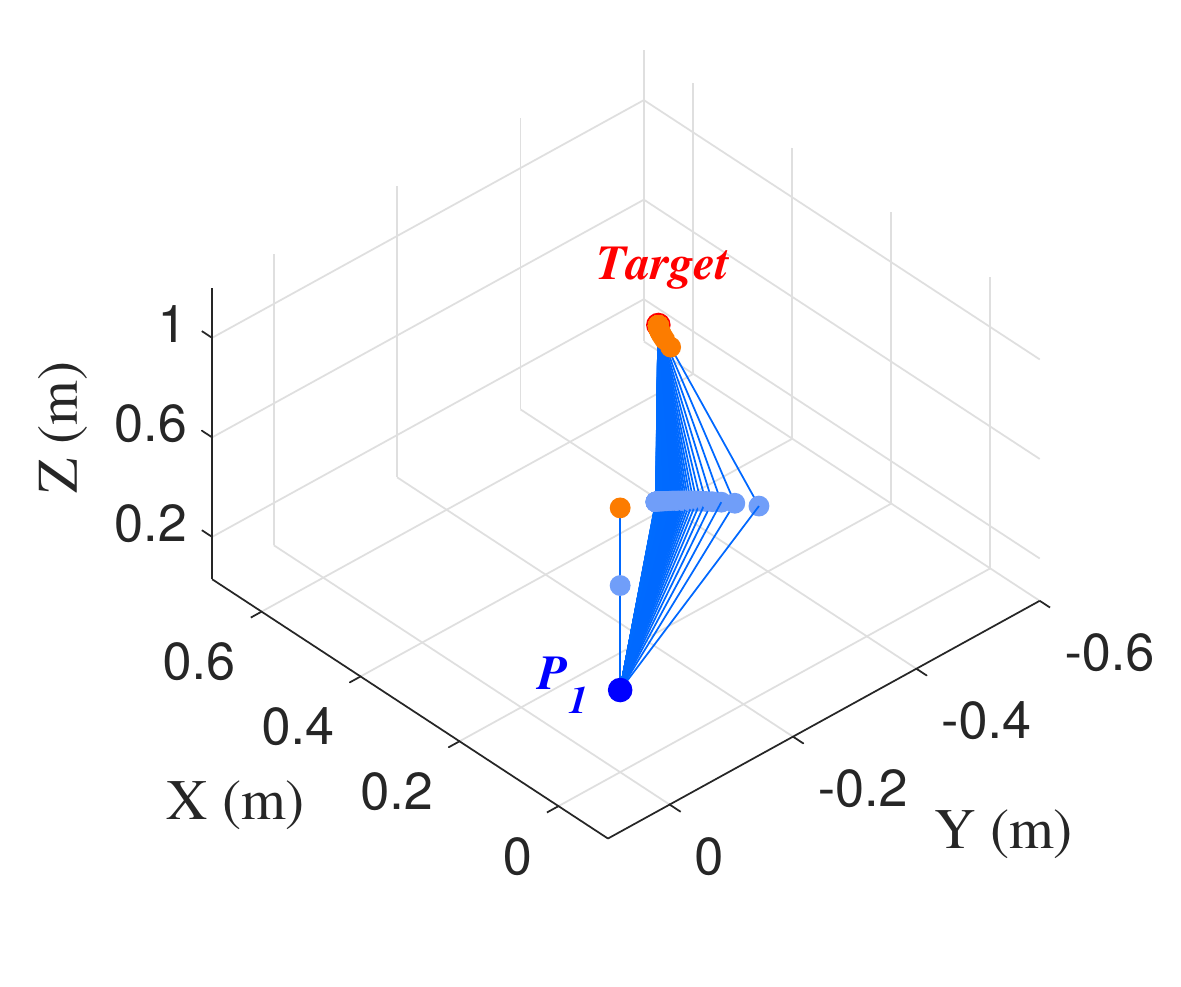}}
	\hspace{0.3cm}
	\subcaptionbox{\label{2d-9}}[4.5cm][c]{\includegraphics[width=4.5cm]{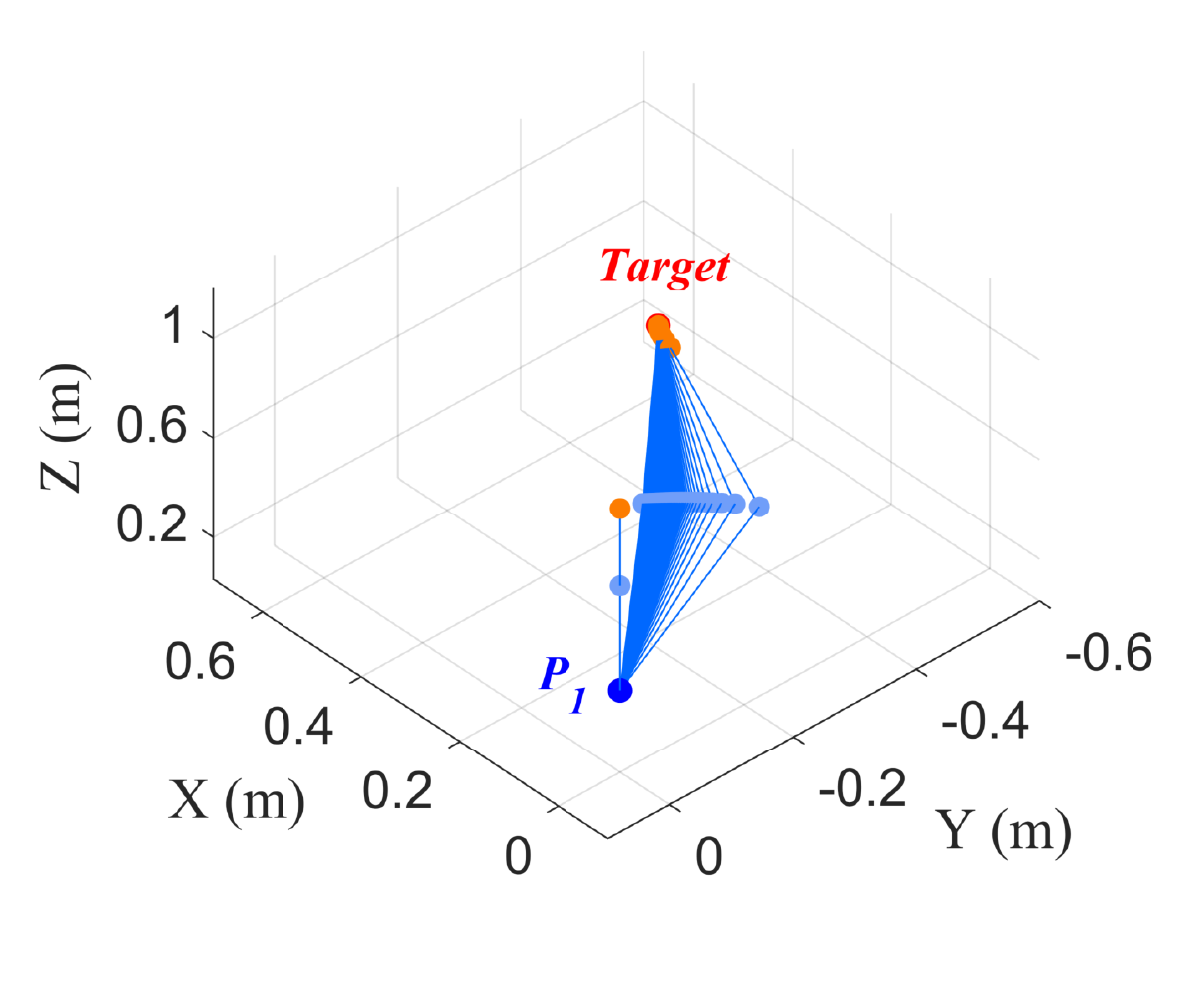}}
	
	\subcaptionbox{\label{2d-10}}[4.5cm][c]{\includegraphics[width=4.5cm]{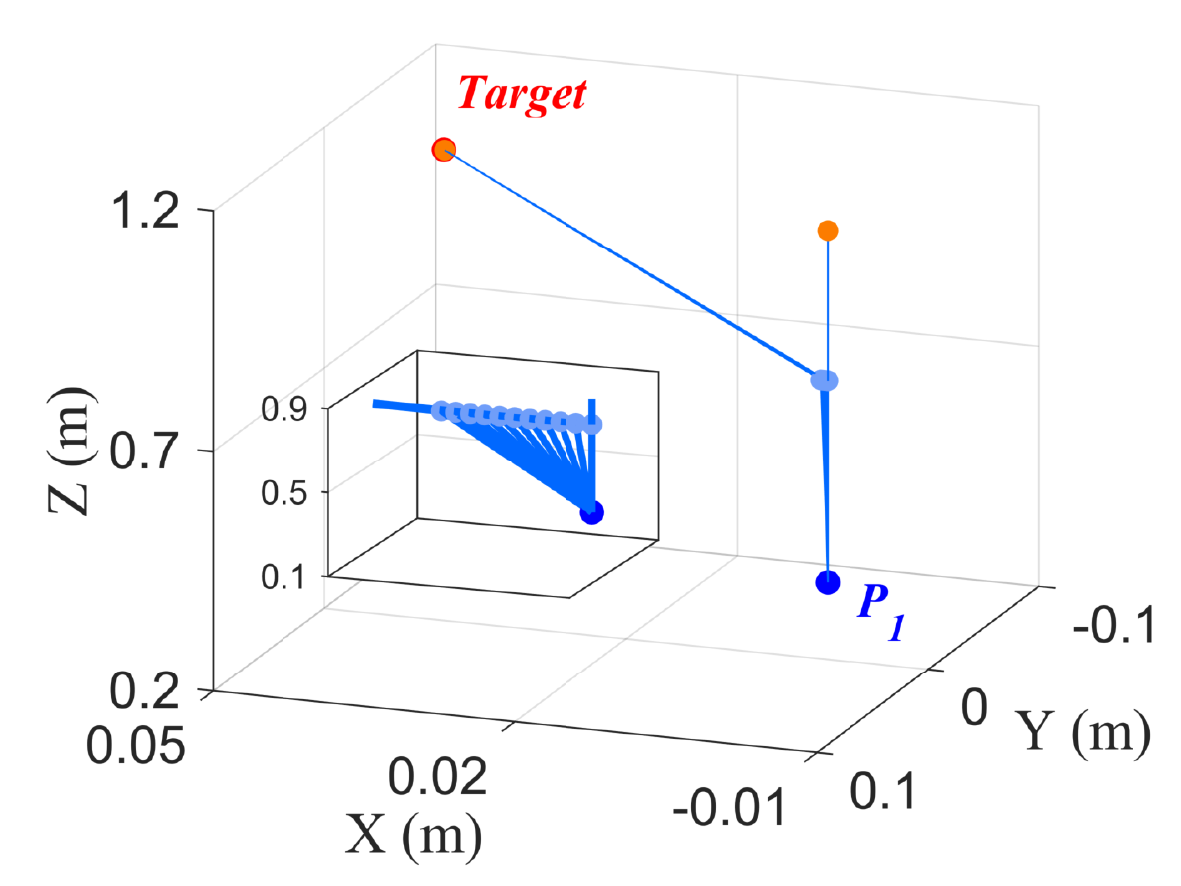}}
	\hspace{0.3cm}
	\subcaptionbox{\label{2d-11}}[4.5cm][c]{\includegraphics[width=4.5cm]{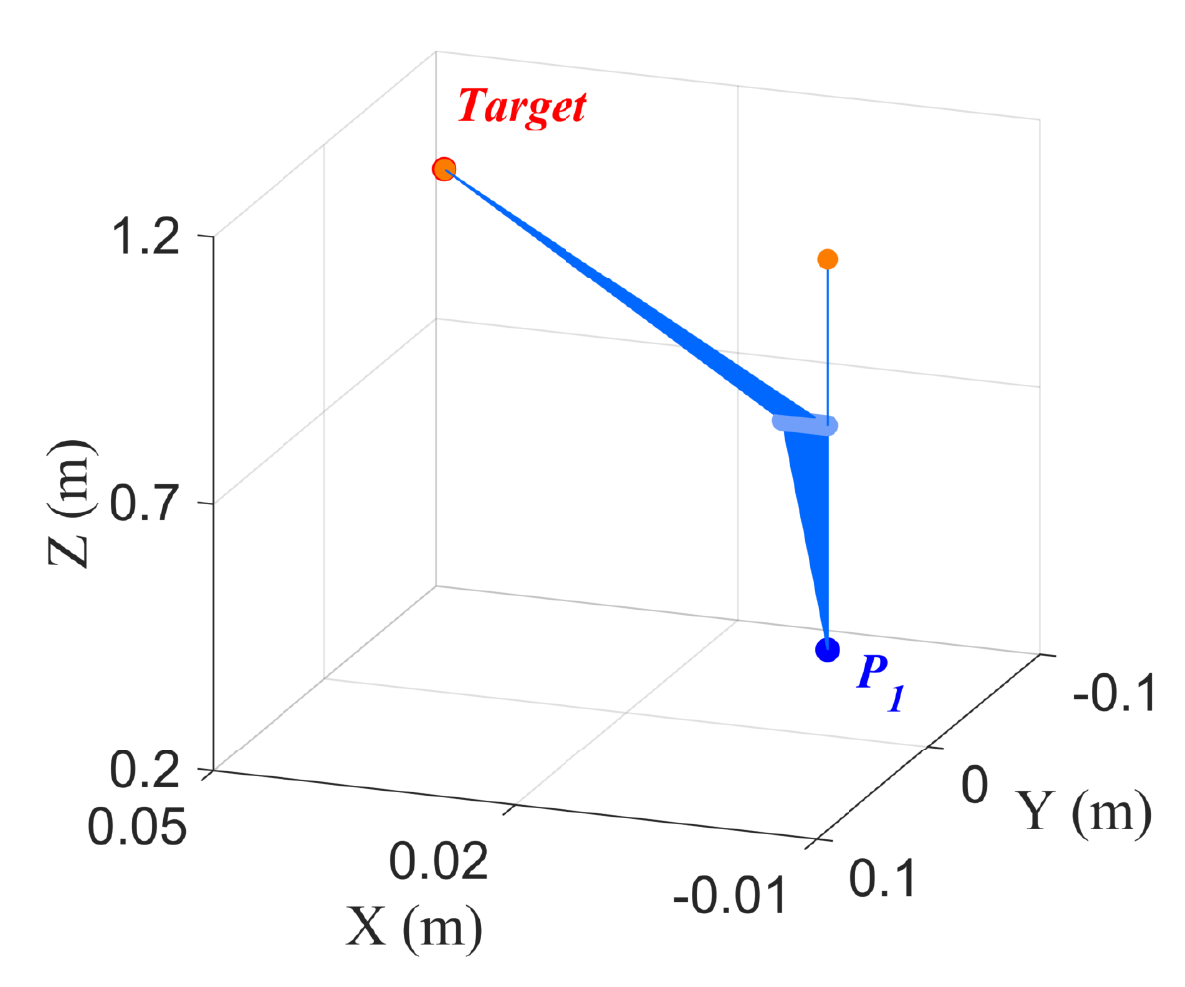}}
	\hspace{0.3cm}
	\subcaptionbox{\label{2d-12}}[4.5cm][c]{\includegraphics[width=4.5cm]{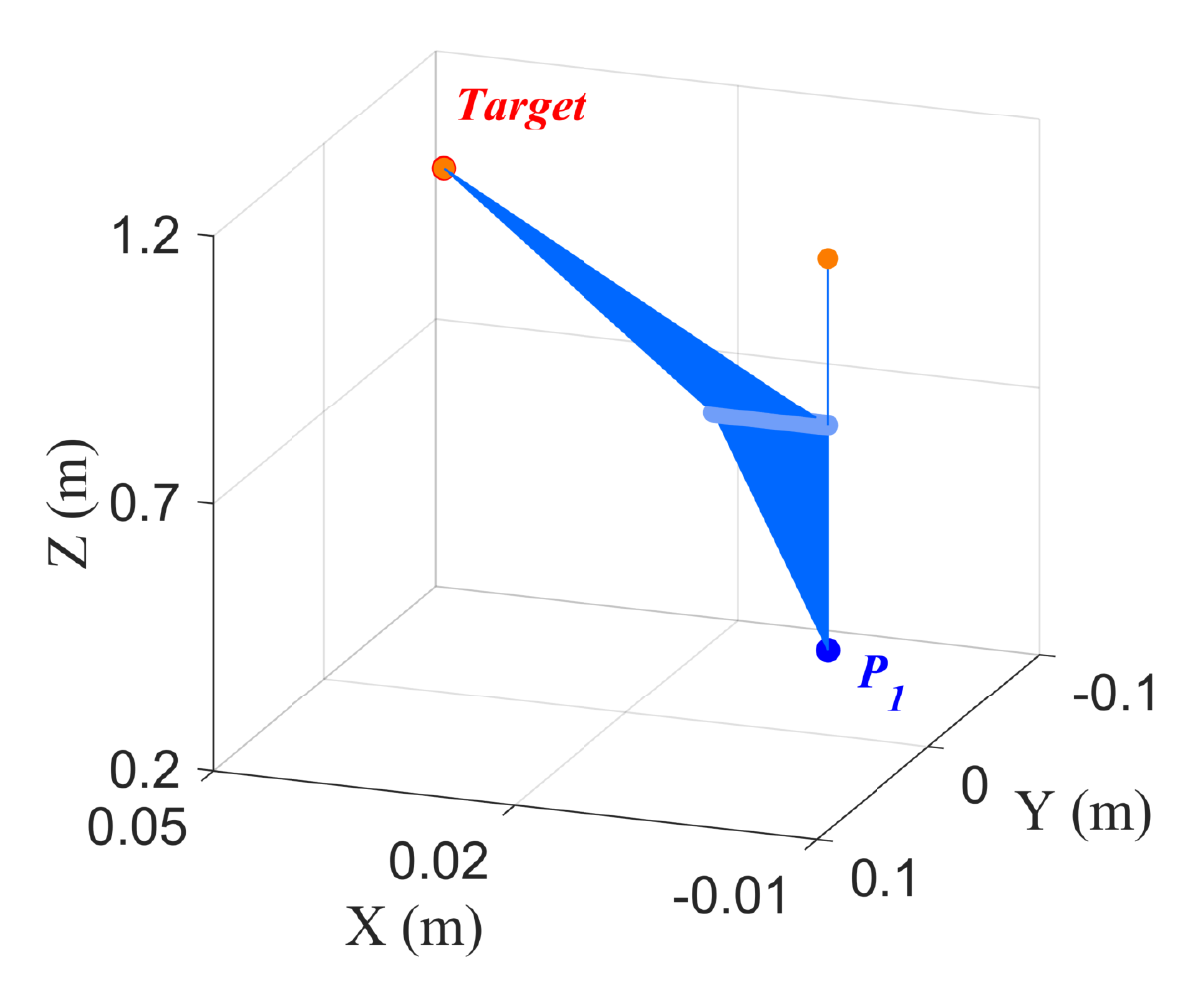}}
	\caption{The iteration processes of the (first two rows) planar 2-DOF and (last two rows) 4-DOF manipulators, where \tikzcircle[blue, fill=blue]{3pt} $ \bm{P}_1 $, \tikzcircle[elbow1_color, fill=elbow1_color]{3pt} $ \bm{P}_2 $, \tikzcircle[elbow2_color, fill=elbow2_color]{3pt} $ \bm{P}_3 $, \tikzcircle{3pt} target, and the initial configurations are indicated. Each column represents a different iteration stage, with iterations 10 and 100 (i.e., $ n = 10 $ and $ n = 100 $) shown in the first two columns. The last column indicates the completion states, i.e., (\subref{2d-3}) $ n = 659 $, (\subref{2d-6}) $ n = 681 $, (\subref{2d-9}) $ n = 10010 $, and (\subref{2d-12}) $ n = 3069 $.}
	\label{2d}
\end{figure*}

\begin{figure}[t]
	\centering
	\captionsetup[subfigure] {skip=-120pt,slc=off,margin={-5pt, 0pt},labelfont=normalfont} 
	\subcaptionbox{\label{dis-1}}[4.2cm][c]{\includegraphics[width=4.2cm]{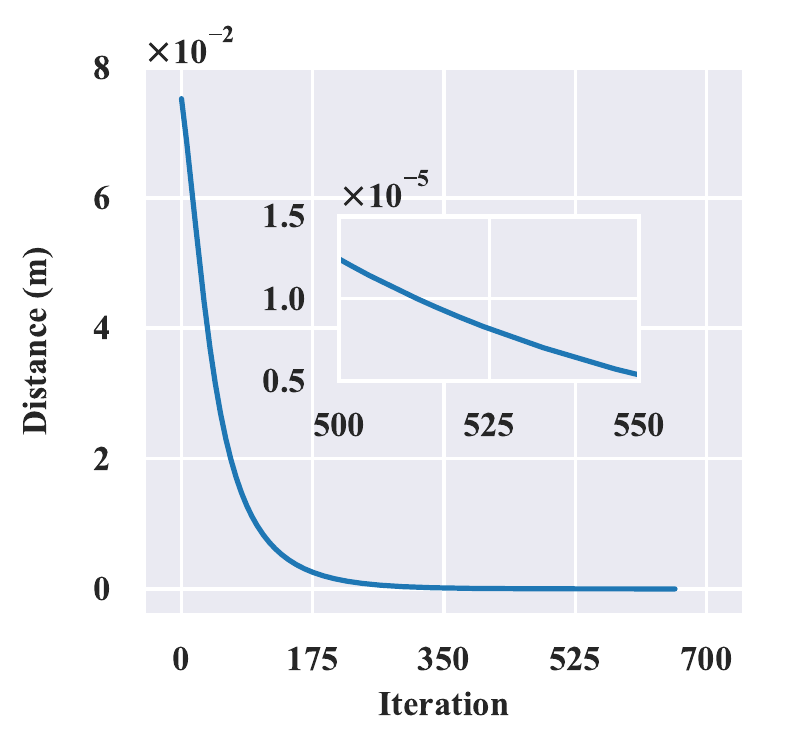}}
	%		\hspace{-0.3cm}
	\subcaptionbox{\label{dis-2}}[4.2cm][c]{\includegraphics[width=4.2cm]{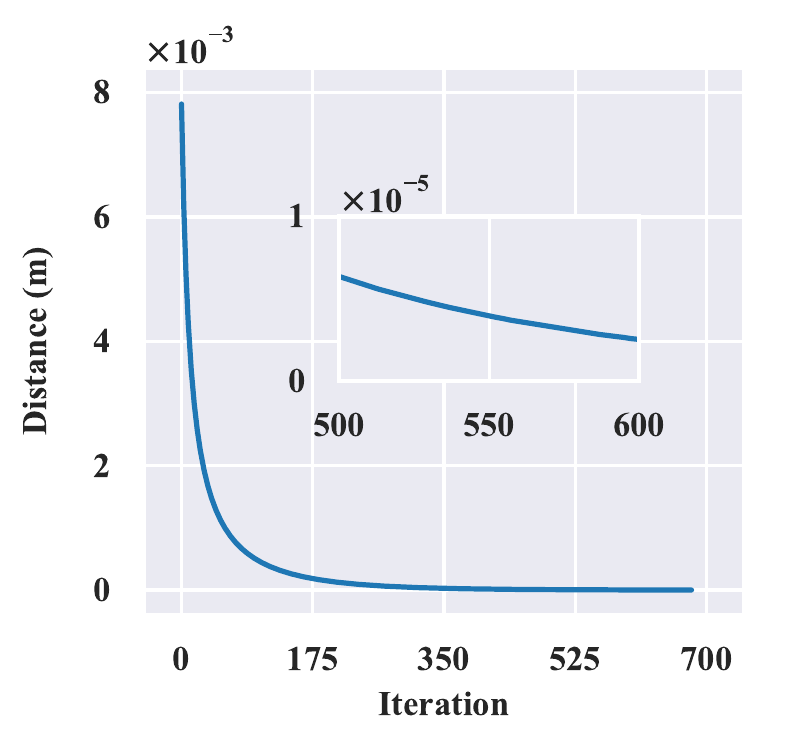}}
	
	\subcaptionbox{\label{dis-3}}[4.2cm][c]{\includegraphics[width=4.2cm]{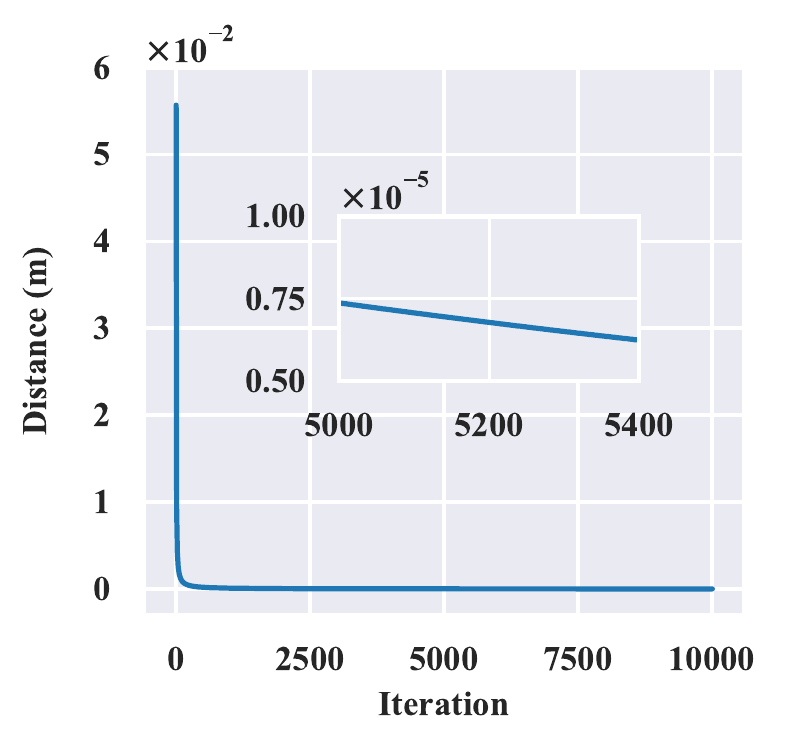}}
	\subcaptionbox{\label{dis-4}}[4.2cm][c]{\includegraphics[width=4.2cm]{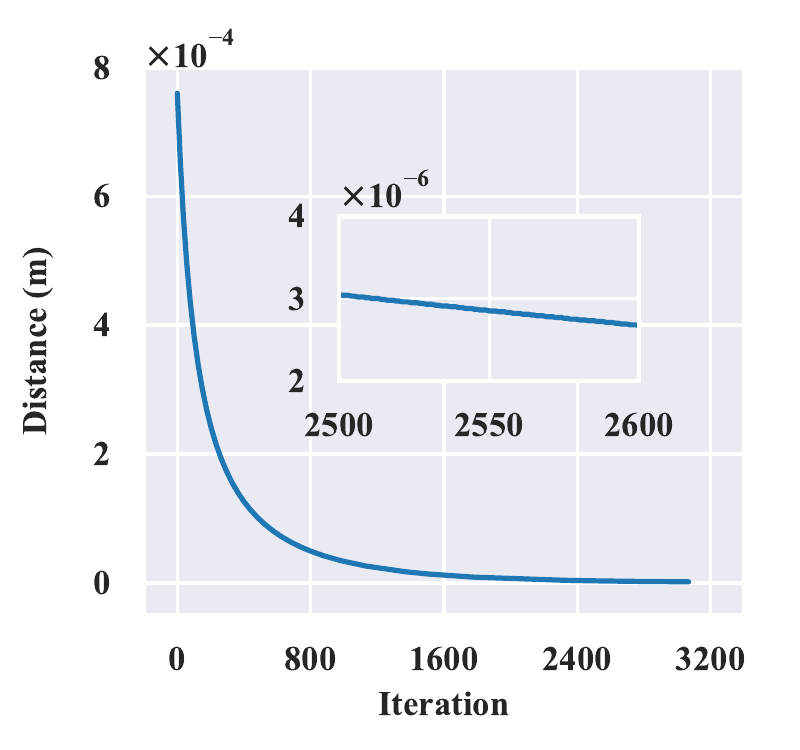}}
	\caption{Variations of the distance between the targets and $ \bm{P}_3 $ of the (first row) 2-DOF and (second row) 4-DOF manipulators during iterations, where (\subref{dis-1}), (\subref{dis-2}), (\subref{dis-3}), and (\subref{dis-4}) correspond to the processes shown in the first to fourth rows in Fig.~\ref{2d}, respectively.}
	\label{distance}
\end{figure}

	As shown in Fig.~\ref{link-1}, for the planar two-link manipulator with a fixed base, the forward phase can be described as
\begin{equation}
	{\bm{S}_1} \buildrel \Delta \over = \left\{ 
	\begin{aligned}
			{\bm{P}_3} &= {\bm{P}_t}\\\
			{\bm{P}_2} &= \left( {1 - {\alpha _3}} \right){\bm{P}_3} + {\alpha _3}{\bm{P}_2}\\
			{\bm{P}_1} &= \left( {1 - {\alpha _2}} \right){\bm{P}_2} + {\alpha _2}{\bm{P}_1},\ if\ \theta _{2_{\min }} \le \varphi_2 \le \theta _{2_{\max }}\\
			{\bm{P}_1} &= \left( {1 - {\alpha _2}{R_{{\bm{z}_2}}}\left( \Delta \varphi_2 \right)} \right){\bm{P}_2} + {\alpha _2}{R_{{{\bm{z}}_2}}}\left( \Delta \varphi_2 \right){\bm{P}_1}\\ 
			& \hspace{5.5cm} else, 
	\end{aligned}
	\right.
	\label{2d_fabrik}
\end{equation}
	where $ {\alpha _i} = {l_i}/{d_i}\ (i = 2, 3)$. $ {l_i} $ is the length of the $ i $-th link, and $ {d_i} $ is the distance between $ {\bm{P}_{i-1}} $ and $ {\bm{P}_{i}} $. $ \varphi_2 $ is the angle between $ \widehat{\bm{P}_3\bm{P}_2} $ and $ \widehat{\bm{P}_2\bm{P}_1} $. $ \bm{z}_2 $ is the rotation axis of the second joint. When $ \varphi_2 $ exceeds the joint limit of $ \theta_2 $, $ \Delta \varphi_2 $ is given by
	\begin{equation}
	\Delta {\varphi_2} = \left\{ \begin{array}{l}
		\theta _{{ {2} }_{\max }} - {\varphi_2}\quad if\ \theta _{{ 2}_{\max }} < {\varphi_2},\\
		\theta _{{ 2 }_{\min }} - {\varphi_2}\quad if\ {\varphi_2} < \theta_{{ 2 }_{\min }}.
	\end{array} \right.
	\label{fai_2}
\end{equation}
$ {\bm{P}_t} $ is the desired iteration target. $ R_{\bm{u}}\left( {\theta} \right)$ represents the rotation around the axis $ \bm{u} $ by an angle $ \theta $ and can be expressed as
\begin{equation}
	R_{\bm{u}} \left( \theta  \right) = \left( {\cos \left( \theta  \right)} \right)I + \left( {\sin \left( \theta  \right)} \right){\left[ \bm{u} \right]_ \times } + \left( {1 - \cos \left( \theta  \right)} \right)\left( {\bm{u} \otimes \bm{u}} \right) ,
	\label{rodrigue}
\end{equation}
where $\left[ \bm{u} \right]_ \times$ is the cross product matrix of $\bm{u}$. Meanwhile, the backward phase can be described as
\begin{equation}
	{\bm{S}_2} \buildrel \Delta \over = \left\{ 
	\begin{aligned}
		{\bm{P}_1} &= \bm{P}_1^{int}\\
		{\bm{P}_i} &= \left( {1 - {\alpha _i}} \right){\bm{P}_{i - 1}} + {\alpha _i}{\bm{P}_i}\left( {i = 2,3} \right)\\
		& \hspace{1.5cm} if\ \theta _{{\left( {i - 1} \right)}_{\min }} \le {\varphi _{i - 1}} \le \theta _{{\left( {i - 1} \right)}_{\max }},\\
		{\bm{P}_i} &= \left( {1 - {\alpha _i}{R_{{\bm{z}_{i - 1}}}}\left( \Delta \varphi _{i - 1} \right)} \right){\bm{P}_{i - 1}} + \\
		& \hspace{1cm} {\alpha _i}{R_{{\bm{z}_{i - 1}}}}\left( \Delta \varphi _{i - 1} \right){\bm{P}_i}\ \left( {i = 2,3} \right)\ else,
	\end{aligned}
	\right.
	\label{2d_fabrik_2}
\end{equation}
where $ \bm{P}_1^{int} $ is the fixed position of $ \bm{P}_1 $. In the backward phase, $ \varphi_{i-1} $ denotes the angle between $ \widehat {\bm{P}_{i-2} \bm{P}_{i-1}} $ and $ \widehat {\bm{P}_{i-1} \bm{P}_{i}} $. Analogously to Eq.~\eqref{fai_2}, $ \Delta \varphi_{i-1} $ can be easily derived when $ \varphi_{i-1} $ exceeds the joint limit of $ {\theta _{i - 1}} $. Notably, for the 4-DOF manipulator in Fig.~\ref{link-2}, the adjustment of $ \bm{P}_2 $ in the backward phase is different from that of the planar manipulator. When $ \varphi_{1} $ exceeds the joint limit of the ball joint, the rotation axis $ \bm{u} $ in $ R_{\bm{u}} \left( \Delta \varphi_{1}  \right) $ is given by
\begin{equation}
	\bm{u} = \widehat {{\bm{P}_0}{\bm{P}_1}} \times \widehat {{\bm{P}_1}{\bm{P}_2}}.
	\label{ball_joint}
\end{equation}
%%%%%%%%%%%%%%%%%%%%%%%%%%%%%%%%%%%%%%%%%%%%%%%%%%%%%%%%%%%%%%%%%%%%
\begin{table}[th!]
	\caption{MATHEMATICAL NOTATIONS}
	\label{notation}
	\begin{center}
		\renewcommand\arraystretch{1.3}  % 修改行距
		\begin{tabular}{C{1.6cm} L{5.9cm}}
			\toprule
			Notation & Meaning \\
			\midrule
			$\bm{S}_{1},\ \bm{S}_{2}$ & Sets of joint Cartesian positions in different phases \\
			$\bm{P}_{i}$ & The Cartesian position of the $i$-th joint \\
			$\bm{P}_{t}$ & Desired iteration target \\
			$ n_l $ & Switch index for FABRIK and the SQP algorithm  \\
			$ n_{max} $ & Iteration limit of FABRIK  \\
			$\bm{v}_{init}$ &  Initial direction vector of the kinematic chain for iteration \\
			$ \widehat {\bm{l}_{id}} $ & Desired unit direction vector of the $ i $-th link \\
			$ \widehat {\bm{x}_{id}} $, $ \widehat {\bm{y}_{id}} $, $ \widehat {\bm{z}_{id}} $ & Desired $ x $, $ y $, and $ z $-axes unit vectors of the $ i $-th frame \\ 
			$ dist $ &  The distance between $\bm{P}_{t}$ and the end of the kinematic chain involved in iteration \\
			$ k $ & The degree of freedom of the manipulator \\
			$\bm{\Theta}_{init}^k $, $\bm{\Theta}_{des}^k $ & The initial and desired joint angle vectors of the $ k $-DOF manipulator\\
			$ \varepsilon_{pos} $, $ \varepsilon_{rot} $ & The position and pose errors of the end-effector \\
			$ \varepsilon_{tol} $ & Cartesian error constraint \\
			$ \bm{\mathcal{S}} $ & Set containing joint angle vectors that satisfy $ \varepsilon_{tol} $ \\
			\bottomrule
		\end{tabular}
	\end{center}
\end{table}
%%%%%%%%%%%%%%%%%%%%%%%%%%%%%%%%%%%%%%%%%%%%%%%%%
\vskip -15pt
The convergence analysis is conducted with the $ \varepsilon_{tol} = 10^{-6}$ Cartesian error constraint. The configurations after each backward phase and the position variations of $ \bm{P}_2 $ and $ \bm{P}_3 $ are shown in Fig.~\ref{2d}. Figs.~\ref{2d-1}-\ref{2d-3} and \ref{2d-7}-\ref{2d-9} illustrate the scenarios in which manipulators must be significantly changed to reach targets. As shown in Fig.~\ref{2d-1}, for the 2-DOF planar manipulator, the initial direction vector for iteration is $ \widehat {{\bm{P}_1}{\bm{P}_2}} $ = $ \widehat {{\bm{P}_2}{\bm{P}_3}}$ = ${\bm{v}_{init}} $ = $\big[$-0.984, 0.178, 0$\big]^T$, the target for $ \bm{P}_3 $ is $ \bm{P}_t $ = $\big[$0.041, -0.007, 0.135$\big]^T$m, and the joint limits are both $ \left[ { - \pi ,\pi } \right] $ rad. $ \bm{P}_3 $ approaches the target extremely fast during the first ten iterations (i.e., $ n = 10 $). The iterative process depicted in Fig.~\ref{2d-3} is visually almost identical to that in Fig.~\ref{2d-2}, but with 559 more iterations, which implies that the subsequent joint position updates are extremely slow in comparison to the initial stage in Fig.~\ref{2d-1}. The entire process takes 0.9 ms, of which 0.765 ms is spent transitioning from Fig.~\ref{2d-2} to Fig.~\ref{2d-3}. Then, for the 4-DOF manipulator, $ \bm{P}_t = \left[0.606, -0.548, 0.290 \right]^T$m and $ {\bm{v}_{init}} = \left[ {0, 0, 1} \right]^T$. The iteration processes in Figs.~\ref{2d-7}-\ref{2d-9} span 10010 iterations and take 179.927 ms. Additionally, when the target is close to the initial position of $ {\bm{P}_3} $, the kinematic chain only needs to bend slightly based on the initial configuration. However, the processes depicted in Figs.~\ref{2d-4}-\ref{2d-6} and \ref{2d-10}-\ref{2d-12} span 681 (19.1 ms) and 3069 (54.3 ms) iterations, respectively. Besides, Fig.~\ref{distance} further depicts a significant decrease in the gradient of the curves of the distance between $ \bm{P}_3 $ and the targets (i.e., $ dist $). In summary, the above phenomena are typically related to the initial configuration and target position, which means that FABRIK may struggle in inefficient iterative processes under the high precision constraint. Extensive tests reveal that the inefficient iterations typically happen when the kinematic chain should make a little or significant bend to reach the target. Thus, as indicated above, FABRIK's fluctuating convergence property prevents it from providing stable real-time motion when applied to manipulators.

%%%%%%%%%%%%%%%%%%%%%%%%%%%%%%%%%%%%%%%%%
\section{Algorithm}
\label{sec3}
In addition to the phenomena summarized in Sec.~\ref{sec2}, FABRIK can still exhibit excellent computation efficiency in most situations. Thus, in order to solve the problem discussed above and apply FABRIK to manipulators, an algorithm combining FABRIK and the SQP algorithm is proposed in this section. Then, this algorithm is applied to the UR5 and KUKA manipulators, which includes the analytical derivation to deduce joint angles.
\subsection{Algorithmic Procedure}
As shown in Fig.~\ref{distance}, the inefficient iterations of FABRIK usually start early in the entire solution process. These iterations can be replaced and formulated as the SQP problem. Hence, the sequential least-squares quadratic programming (SLSQP) algorithm \cite{kraft1988software}, which is updated by the BFGS algorithm and provides a near-quadratic convergence rate, is a suitable choice for SQP. However, the switch principle of FABRIK and SLSQP cannot be determined by judging the value of $dist$ in each loop and its difference between two consecutive iterations because $ dist $ varies in different steps at the initial stage of iteration for different targets, as shown in Fig.~\ref{distance}. Thus, a switch index (i.e., $ n_l$) for FABRIK and SLSQP, which is also an iteration limit for FABRIK, can be specified generally. The optimization phase will be omitted if FABRIK can converge within $ n_l $. After iterations or optimization, a link configuration that satisfies $ \varepsilon _{tol} $ can be obtained. Furthermore, in order to deduce joint angles from joint positions and link directions, some necessary analytical processes that depend on manipulator structures are required. Based on the designed combination, which is summarized in Fig.~\ref{all_procedure}, this algorithm will be applied to the UR5 and KUKA manipulators and explained in detail in the following subsections. 
\subsection{UR5 Manipulator}
\label{UR5_section}
\begin{figure}[t]
	\centering
	\captionsetup[subfigure] {skip=0pt,slc=off,margin={55pt, 0pt},labelfont=normalfont} 
	\subcaptionbox{\label{ur5}}{\includegraphics[scale=.5]{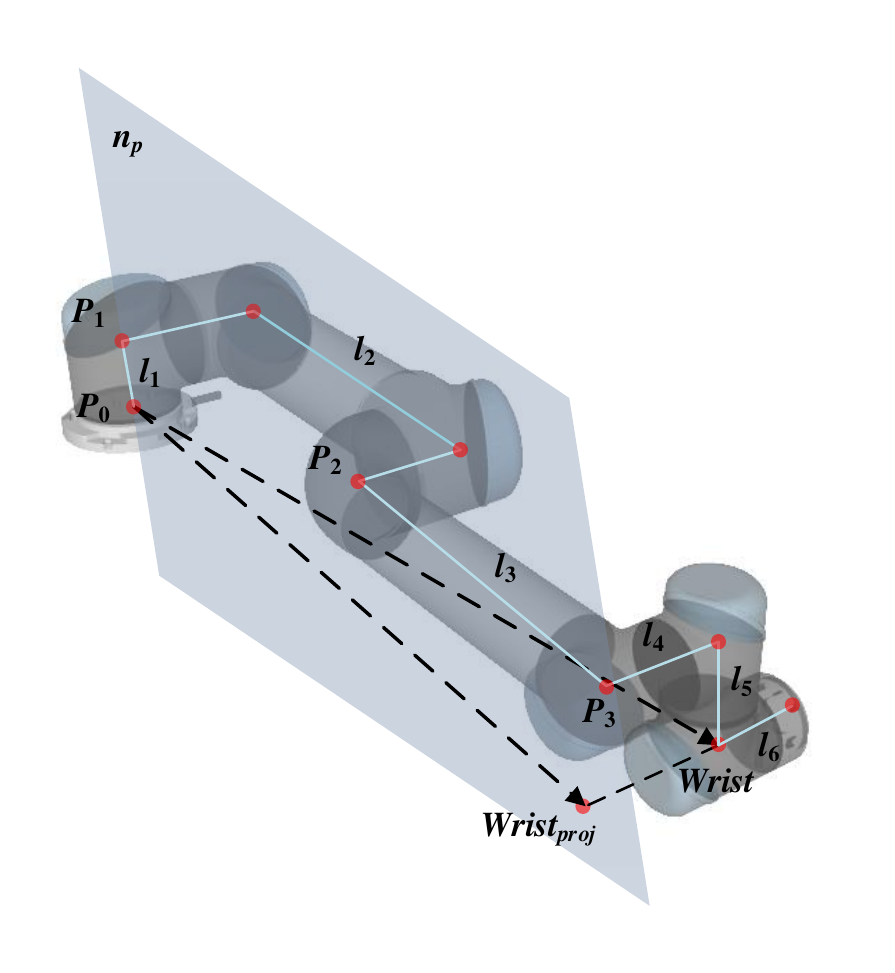}}
	%	\hspace{0.5cm}
	\subcaptionbox{\label{kuka}}{\includegraphics[scale=.5]{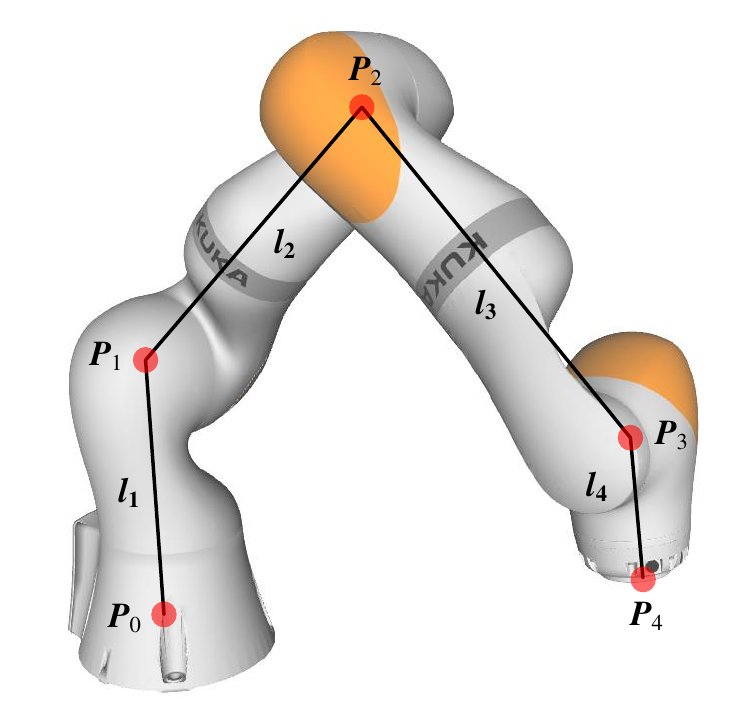} \vspace{0.55cm}}
	%	\vskip -7pt
	\caption{Illustrations of the (\subref{ur5}) UR5 and (\subref{kuka}) KUKA kinematic chains}
	\label{ur5_kuka_chain}
\end{figure}

The 2-D iterations of FABRIK can be implemented on the UR5 manipulator with some analytical derivations. The DH parameters of UR5 can be found in \cite{andersen2018kinematics}. As shown in Fig.~\ref{ur5}, links $ l_2 $ and $ l_3 $ are involved in the iterations that happen in plane $ \bm{n_p} $. First, the wrist position in the base frame is given by
\begin{equation}
	\label{UR5-1}
	^0\bm{P}_w = {\bm{P}_{des}} - \underbrace {{R_{des}}{{\left[ {0,0, {l_6}} \right]}^T}}_{\widehat {{\bm{l}_{6d}}}},
\end{equation}
where $ {\bm{P}}_{ des} $ and $ {R}_{ des} $ are the position vector and rotation matrix of the desired homogeneous matrix (i.e., $ {T}_{des}^k $, where $ k $ is the DOF of the current manipulator). $ \widehat {\bm{l}_{id}} \left( {i = 1,2,...,6} \right)$ is the desired unit direction vector of the $ i $-th link. $ \theta_1 $ is first obtained by
\begin{equation}
	\label{UR5-2}
	\begin{aligned}
		{\theta _1} = \pm {\rm{acos}}\left( {\frac{{{l_4}}}{{\sqrt {{}^0{\bm{P}_{w_x}}^2 + {}^0{\bm{P}_{w_y}}^2} }}} \right) + \frac{\pi }{2} + \\
		{\rm{atan}}2\left( {{}^0{\bm{P}_{w_y}},{}^0{\bm{P}_{w_x}}} \right),
	\end{aligned}
\end{equation}
where ${{}^0{\bm{P}_{w_x}}}$ and ${{}^0{\bm{P}_{w_y}}}$ are the $ x $, $ y $ coordinates of $ {}^0{\bm{P}_w} $. The position of $ Wrist_{proj} $ (i.e., $ {\bm{P}_{{w_{proj}}}} $) can be obtained by projecting the base-to-wrist vector onto $ \bm{n_p} $. $ \widehat {\bm{l}_{5d}} $ can then be deduced by calculating
\begin{equation}
	\label{UR5-3}
	\widehat {{\bm{l}_{5d}}} =  \pm \left( {\widehat {{\bm{z}_{2d}}}  \times \widehat {{\bm{l}_{6d}}}} \right),
\end{equation}
in which $ \widehat {{\bm{z}_{2d}}} = \left[ {\sin {\theta _1}, - \cos {\theta _1},0} \right]^T$ is the desired $z$-axis unit vector of the second frame. The iteration target can be given by 
\begin{equation}
	\label{ur5_target}
	\bm{P}_t = \bm{P}_{{w_{proj}}} - \widehat {{\bm{l}_{5d}}} \times {l_5}.
\end{equation}
%%%%%%%%%%%%%%%%%%%%%%%%%%%
%%%%%%%%%%%%%%%%%%%%%%%%%%%

$ \theta_5 $ and $ \theta_6 $ can be set to zero when $ \widehat {\bm{z}_{2d}} $ and $ \widehat {\bm{l}_{6d}} $ are collinear. The initial direction vector of the kinematic chain for iteration is given by
\begin{equation}
	\bm{v}_{init} = -\widehat{{\bm{x}_{1d}}} = \left[ { - \cos \left( {{\theta _1}} \right), - \sin \left( {{\theta _1}} \right),0} \right]^T,
	\label{ur5_init}
\end{equation} 
in which $ \widehat{{\bm{x}_{id}}} $ is the desired $ x $-axis unit vector of the $ i $-th ($1 \le i \le 6$) frame. Eqs.~\eqref{2d_fabrik} and \eqref{2d_fabrik_2} are iterated when the target is reachable $ \big($i.e., $ {\left\| {\overrightarrow {{\bm{P}_t}{\bm{P}_1}} } \right\|_2} \le {l_2} + {l_3} \big)$. If $ dist \le \varepsilon_{tol}$ and $ n \le n_l $ are satisfied, the desired joint angle vector $ \bm{\Theta}_{des}^6 $ can be deduced without performing optimization. $ \theta_2 $, $ \theta_3 $, and $ \theta_4 $ can be first obtained by 
\begin{equation}
	\label{UR5-5}
	\begin{aligned}
		{\theta _2} = sign\left\langle {\widehat {{\bm{z}_{2d}}} , - \widehat {{\bm{x}_{1d}}}  \times \widehat {{\bm{l}_{2d}}}} \right\rangle  \cdot \arccos {\left\langle -{\widehat {{\bm{x}_{1d}}},\widehat {{\bm{l}_{2d}}}} \right\rangle }, 
	\end{aligned}
\end{equation}
\begin{equation}
	\label{theta3}
	{\theta _3} = sign\left\langle {\widehat {{\bm{z}_{2d}}} ,\widehat {{\bm{l}_{2d}}} \times \widehat {{\bm{l}_{3d}}}} \right\rangle  \cdot \arccos {\left\langle {\widehat {{\bm{l}_{2d}}},\widehat {{\bm{l}_{3d}}}} \right\rangle },
\end{equation}
and
\begin{equation}
	\label{theta_4}
	{\theta _4} = sign\left\langle {\widehat {{\bm{z}_{2d}}} ,\widehat {{\bm{l}_{3d}}} \times \widehat {\bm{l}_{5d}}} \right\rangle  \cdot \arccos {\left\langle {\widehat {{\bm{l}_{3d}}},\widehat {{\bm{l}_{5d}}}} \right\rangle } - \frac{\pi }{2}.
\end{equation}
$ \theta_5 $ is then given by 
\begin{equation}
	\label{UR5-8}
	{\theta _5} = sign\left\langle {\widehat {{\bm{l}_{5d}}} , \widehat {\bm{y}_{5d}} } \times \widehat {\bm{l}_{6d}}  \right\rangle  \cdot \arccos  \left\langle { \widehat {\bm{y}_{5d}}, \widehat {\bm{l}_{6d}} } \right\rangle,
\end{equation}
where $ {\theta _5} $ should be presumed to be zero to obtain $ \widehat {\bm{y}_{5d}} = \widehat {{\bm{z}_{2d}}}$. Given $ \theta_5$,  $ \theta_6 $ can be calculated by
\begin{equation}
	\label{UR5-11}
	{\theta _6} = sign\left\langle {\widehat {{\bm{z}_{6d}}} ,\widehat {{\bm{x}_{5d}}}  \times \widehat {{\bm{x}_{6d}}} } \right\rangle  \cdot \arccos  \left\langle  \widehat {{\bm{x}_{5d}}}, \widehat {{\bm{x}_{6d}}}  \right\rangle,
\end{equation}
in which $ \widehat {\bm{x}_{5d}} $ is the first three rows of the first column of $ \prod\limits_{n = 1}^5 {{A_n}} $. $ \widehat {\bm{x}_{6d}} $ and $ \widehat {\bm{z}_{6d}} $ are the $ x $ and $ z $-column vectors of $ R_{des} $. Since the sign of $ \widehat {{\bm{l}_{5d}}} $ and $ \theta_1 $ cannot be determined during calculation, each of the derived joint angle vectors should be filtered with the following equation:
\begin{equation}
	\mathcal{D} \left( \bm{\Theta}_{temp}^{k} \right) = {\varepsilon _{ rot}} + {\varepsilon _{ pos}},
	\label{pos_rot}
\end{equation}
where 
\begin{equation}
	\begin{aligned}
		{\varepsilon _{ rot}} &= \arccos \left( {\frac{{tr\left( {R_{ temp}^{-1}{R_{ des}}} \right) - 1}}{2}} \right)\ \textrm{and}\\ 
		{\varepsilon _{ pos}} &= {\left\| {{{\bm{P}}_{ temp}} - {{\bm{P}}_{ des}}} \right\|_{ 2}}. \nonumber
	\end{aligned}
	\label{error_pos_rot}
\end{equation}
$ {\varepsilon _{ rot}} $ and $ {\varepsilon _{ pos}} $ are the pose and position errors induced by the current joint angle vector $ {\bm{\Theta}_{temp}^k} \in \mathbb{R}^k $. $ R_{temp} $ and $ \bm{P}_{temp} $ are the rotation matrix and position vector of $ T_{temp} $, which can be obtained by 
\begin{equation}
	{T_{temp}}\left( \bm{\Theta}_{temp}^k  \right) = \prod\limits_{n = 1}^k {{A_n}}.
	\label{eq10}
\end{equation}
The set of joint angle vectors satisfying $ \varepsilon _{tol} $ is given by 
\begin{equation}
	\bm{\mathcal{S}} \buildrel \Delta \over = \left\{ {\bm{\Theta}^k : \mathcal{D} \left( \bm{\Theta}^k  \right) \le {\varepsilon_{tol}}} \right\}.
	\label{desired_set}
\end{equation}
The desired joint angle vector $ \bm{\Theta}_{des}^k $ can then be obtained by
\begin{equation}
	\bm{\Theta}_{des}^k = \mathop {\arg \min }\limits_{\bm{\Theta}^k \in \bm{\mathcal{S}}} {\left\| \bm{\Theta}^k - \bm{\Theta}_{init}^k \right\|_1},
	\label{eq11}
\end{equation}
where $ \bm{\Theta}_{init}^k $ is the initial joint angle vector. The flows indicated by the blue-gray arrows in Fig.~\ref{all_procedure} are implemented by the above processes. Moreover, when $ n = n_l $ and $ dist > {\varepsilon _{tol}} $, the current values of joints 2 and 3 $ \big( \textrm{i.e.,}\ {\theta_{i_{seed}}},\ \textrm{for}\ {i = 2, 3} \big)$ should be used as the initial values of SLSQP. The optimization phase, which is represented by the black arrows in Fig.~\ref{all_procedure}, is then triggered and formulated as 
\begin{equation}
	\begin{aligned}
		\min \big[ {\big( {\bm{P}_{{3_x}}} - \bm{P}_{{t_x}} \big)}^2 &+ {{\left( {{\bm{P}_{{3_y}}} - {\bm{P}_{{t_y}}}} \right)}^2} + {{\left( {{\bm{P}_{{3_z}}} - {\bm{P}_{{t_z}}}} \right)}^2} \big] \\
		{\rm{s.t.}}\ &{\theta _{{i_{\min }}}} \le {\theta _i} \le {\theta _{{i_{\max }}}},
	\end{aligned}
	\label{opt_ur5}
\end{equation}
where 
\begin{equation}
	\begin{aligned}
		{\bm{P}_{{3_x}}} &= \cos \left( {{\theta _1}} \right)\left( { - \left( {{l_2}\cos \left( {{\theta _2}} \right) + {l_2}\cos \left( {{\theta _2} + {\theta _3}} \right)} \right)} \right), \nonumber \\
		{\bm{P}_{{3_y}}} &= \sin \left( {{\theta _1}} \right)\left( { - \left( {{l_2}\cos \left( {{\theta _2}} \right) + {l_2}\cos \left( {{\theta _2} + {\theta _3}} \right)} \right)} \right),\ \textrm{and} \\ 
		{\bm{P}_{{3_z}}} &= {l_1} - \left( {{l_2}\sin \left( {{\theta _2}} \right) + {l_3}\sin \left( {{\theta _2} + {\theta _3}} \right)} \right). \\
	\end{aligned}
	\label{opt_ur5_1_3}
\end{equation}
$  {\varepsilon _{tol}}^2 $ is set as the stopping criteria. $\widehat {\bm{l}_{2d}} $ and $ \widehat {\bm{l}_{3d}} $ can be easily deduced after optimizing the objective function. Eqs.~\eqref{theta_4}-\eqref{eq11} are then used to deduce $ \bm{\Theta}_{des}^6 $. The above implementation is given in Algorithm \ref{alg_ur5}. 

\begin{algorithm}[t] % 显示行号
	\caption{: Solution to the UR5 manipulator using the combined algorithm}
	\label{alg_ur5}
	\begin{algorithmic}[1]
		\Require {${\bm{\Theta}}_{init}^6$, ${T_{des}^6}$, and $\varepsilon_{tol}$};
		\Ensure ${\bm{\Theta}} _{des}^6$;
		\State $ ^0\bm{P}_w \leftarrow $ Eq.~\eqref{UR5-1}
		\State $ \bm{P}_t, \bm{v}_{init} \leftarrow $ Eqs.~\eqref{UR5-2}-\eqref{ur5_init}
		\If{the target is reachable}
		\While{$ n \le {n_l} $ and $ dist  > {\varepsilon _{tol}}$}
		\State Eqs.~\eqref{2d_fabrik} and \eqref{2d_fabrik_2}
		\EndWhile
		\If{$ n \le {n_l} $ and $ dist \le {\varepsilon _{tol}}$}
		\State $ {\bm{\Theta}} _{des}^6 \leftarrow $  Eqs.~\eqref{UR5-5}-\eqref{eq11}
		\ElsIf{$ n = {n_l} $ and $ dist  > {\varepsilon _{tol}}$}
		\State $  \widehat {\bm{l}_{2d}} ,\widehat {{\bm{l}_{3d}}} \leftarrow $ Eq.~\eqref{opt_ur5}
		\State $ {\bm{\Theta}} _{des}^6 \leftarrow $  Eqs.~\eqref{theta_4}-\eqref{eq11}
		\EndIf
		\EndIf
	\end{algorithmic}
\end{algorithm}
%%%%%%%%%%%%%%%%%%%%%%%%%%%%%%%%%%%%%%%
\subsection{KUKA Manipulator}
Under the pose constraint, the links of the KUKA manipulator that actually participate in the iterations can be simplified to $ l_2 $ and $ l_3 $ in Fig.~\ref{kuka} using
\begin{equation}
	\bm{P}_t = {\bm{P}_{des}} - \underbrace {{R_{des}}{{\left[ 0, 0, l_4 \right]}^T}}_{\widehat {{\bm{l}_{4d}}} },
	\label{eq_target}
\end{equation}
where $ {\widehat {\bm{l}_{4d}}} $ is the desired direction vector of the last link. After that, only four DOFs from shoulder to elbow, which determine the wrist position, should be taken into consideration, and the EE pose can be guaranteed. Eqs.~\eqref{2d_fabrik} and \eqref{2d_fabrik_2} are iterated based on $ \bm{v}_{init} $. If $ n \le {n_l} $ and $ dist \le {\varepsilon _{tol}} $ are satisfied, the desired link direction vectors $ \left\{ \widehat {{\bm{l}_{2d}}} ,\widehat {{\bm{l}_{3d}}} ,\widehat {{\bm{l}_{4d}}}  \right\} $ and joint positions $ \left\{ {{\bm{P}_2},{\bm{P}_3},{\bm{P}_4}} \right\} $ can be obtained and used to derive the desired joint angle vector. $\left| {{\theta _{i}}} \right| \left( {i = 2,4,6} \right)$ are first calculated by
\begin{equation}
	\begin{aligned}
		\left| {{\theta _i}} \right| = \left| {\pi  - \arccos \left( {\frac{{l_{j}^{ 2} + l_{ j + 1}^{ 2} - {{\left( {{{\left\| {\bm{P}_{ j + 1}} - {\bm{P}}_{j-1} \right\|}}} \right)}^{ 2}}}}{{2{l_{ j}}{l_{j + 1}}}}} \right)} \right|\\
		 \left( j = i / 2 \right) .
	\end{aligned}
	\label{eq246}
\end{equation}
$ {\theta _1} $ is then calculated by solving 
\begin{equation}
	{\left( {\prod\limits_{n = 1}^3 {{A_n}} } \right)_{1,4}} = {{\bm{P}}_{{2}_x}},
	\label{eq2}
\end{equation}
where the subscripts $ i $, $ j $ denote the $ i $-th row and $ j $-th column of the corresponding matrix. $ x $ is the $ x $ coordinate of the corresponding position vector. Given $ {\theta _1} $,  $ {\theta _2} $, and $ {\theta _4} $, $ {\theta_3} $ is the solution to the following equation:
\begin{equation}
	{\left( {\prod\limits_{n = 3}^5 {{A_n}} } \right)_{1,4}} = {\left( {\prod\limits_{n = 2}^1 {A_n^{ - 1}} {{\left[ {{\bm{P}_3},1} \right]}^T}} \right)_{1,1}}.
	\label{eq4}
\end{equation}

Using $ {\theta _{i}}  \left( {i = 1,2,3,4,6} \right) $, $ {\theta _{5}} $ is given by 
\begin{equation}
	{\left( {\prod\limits_{n = 5}^7 {{A_n}} } \right)_{1,4}} = {\left( {\prod\limits_{n = 4}^1 {A_n^{ - 1}} T_{des}^7} \right)_{1,4}}.
	\label{eq6}
\end{equation}

Finally, $ \theta_7 $, which is the angle between $\widehat {\bm{x}_{6}} $ and $\widehat {\bm{x}_{7}} $, is given by
\begin{equation}
	\left| {{\theta _{ 7}}} \right| = \left| {\arccos \left\langle  {\widehat {\bm{x}_{ 6}}, \widehat {\bm{x}_{7}}} \right\rangle} \right|,
	\label{eq8}
\end{equation}
where ${\widehat {\bm{x}_{ 6}}}$  and  $\widehat {\bm{x}_{ 7}} $ are the $ x $ column vectors of  $\mathop \prod \limits_{ n = 1}^{ 6} {A_{ n}}$  and $T_{des}^7$, respectively. Finally, $ \bm{\Theta}_{ des}^7$ can be derived by executing Eqs.~\eqref{pos_rot}-\eqref{eq11}. 

When $ n = n_l $ and $ dist > {\varepsilon _{tol}} $, the current values of joints 1, 2, 3, and 4 $ \big(\textrm{i.e.,\ } {\theta _{i_{seed}}},\ \textrm{for}\ {i = 1,2,3,4} \big)$, which will be used as the initial values of SLSQP, should be calculated first using the current joint positions and Eqs.~\eqref{eq246}-\eqref{eq4}. The analytical expression of the wrist position is
\begin{equation}
	{\bm{P}_3} = {\left( {\prod\limits_{n = 1}^5 {{A_n}} } \right)_{\bm{P}}},
	\label{wirst_ana}
\end{equation}
where the subscript $ \bm{P} $ denotes the position vector of the corresponding matrix. Given $ {\theta _{i_{seed}}}\left( {i = 1,2,3,4} \right) $, the objective function $ f $ for minimizing the distance between $ {\bm{P}_3} $ and $ {\bm{P}_t} $ is 
\begin{equation}
	\begin{aligned}
		\min\ & f  \buildrel \Delta \over = {\left\| {{\bm{P}_3} - {\bm{P}_t}} \right\|_2^2} \\
		{\rm{s.t.}}\ &{\theta _{{i_{\min }}}} \le {\theta _i} \le {\theta _{{i_{\max }}}},
	\end{aligned}
	\label{opt}
\end{equation}
where the gradient $ \big($i.e., $ \partial f/\partial {\theta _i},\ \textrm{for}\ 1 \le i \le 4 \big) $ can be represented by $ \theta_1 $, $ \theta_2 $, $ \theta_3 $, and $ \theta_4 $. However, the optimized joint angles are not the desired joint angles because they are obtained without taking joints 5, 6, and 7 into account. Thus, the optimized link direction vectors $ \big($i.e., $ \widehat {{\bm{l}_{2d}}}, \widehat {{\bm{l}_{3d}}} \big) $ should be derived first, and Eqs.~\eqref{eq246}-\eqref{eq8} and \eqref{pos_rot}-\eqref{eq11} are then recalculated to obtain $\bm{\Theta}_{des}^7$. The implementation of the combined algorithm on the KUKA manipulator is shown in Algorithm \ref{alg_1}. 
%%%%%%%%%%%%%%%%%%%%%%%%%%%%%%%%%%%%%%%%%%%%%%%%%%%%%%%
\begin{algorithm}[t] % 显示行号
	\caption{: Solution to the KUKA manipulator using the combined algorithm }
	\label{alg_1}
	\begin{algorithmic}[1]
		\Require {${\bm{\Theta}} _{init}^7$, ${T_{ des}^7}$, $ {\bm{v}_{init}} $, and $\varepsilon _{ tol}$};
		\Ensure ${\bm{\Theta}} _{des}^7$;
		\State $ {{\bm{P}}_{t}} \leftarrow $ Eq.~\eqref{eq_target}
		\If{the target is reachable}
		\While{$ n \le {n_l} $ and $ dist  > {\varepsilon _{tol}}$}
		\State Eqs.~\eqref{2d_fabrik} and \eqref{2d_fabrik_2}
		\EndWhile
		\If{$ n \le {n_l} $ and $ dist \le {\varepsilon _{tol}}$}
		\State $ {\bm{\Theta}}_{des}^7 \leftarrow $  Eqs.~\eqref{eq246}-\eqref{eq8} and  \eqref{pos_rot}-\eqref{eq11}
		\ElsIf{$ n = {n_l} $ and $ dist  > {\varepsilon _{tol}}$}
		\State $  \widehat {{\bm{l}_{2d}}} ,\widehat {{\bm{l}_{3d}}}  \leftarrow $ Eqs.~\eqref{eq246}-\eqref{eq4} and \eqref{opt}
		\State $ {\bm{\Theta}} _{des}^7 \leftarrow $  Eqs.~\eqref{eq246}-\eqref{eq8} and \eqref{pos_rot}-\eqref{eq11}
		\EndIf
		\EndIf
	\end{algorithmic}
\end{algorithm}
%%%%%%%%%%%%%%%%%%%%%%%%%%%%%%%%%%%%%%%%%%%%%%%%%%%%%%%%%%%%%%%%%%%%%%%%%%%%%%%%
\section{Experiments}
\label{sec4}
This section begins with the convergence comparison of the combined algorithm and FABRIK, then quantitatively tests them with 10,000 random IK queries, which are constructed by reachable configurations, and finally applies the combined algorithm to the UR5 and KUKA manipulators to provide real-time motions.
\begin{figure*}[t]
	\captionsetup[subfigure] {skip=-120pt,slc=off,margin={0pt, 0pt},labelfont=normalfont} 
	\centering
	\subcaptionbox{\label{comp_1}}[5.5cm][c]{\includegraphics[width=5.5cm]{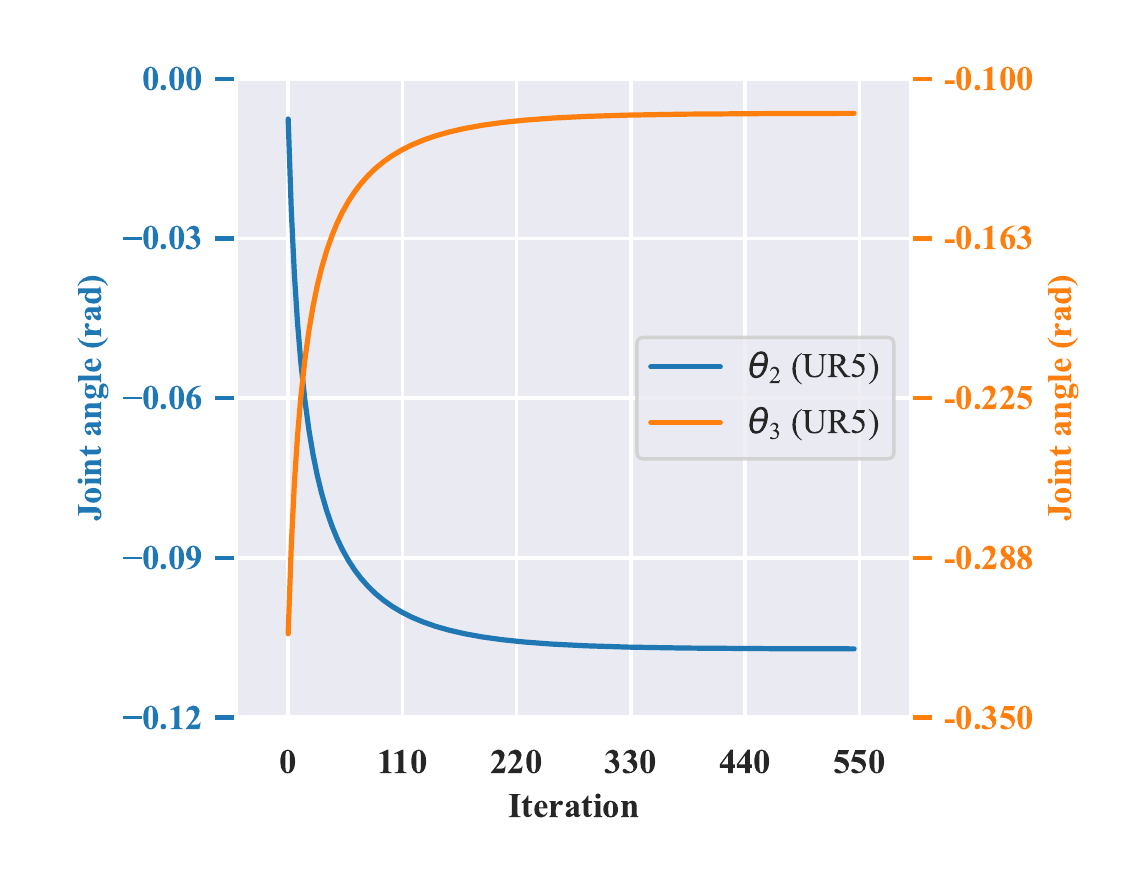}}
	\hspace{-0.3cm}
	\subcaptionbox{\label{comp_2}}[5.5cm][c]{\includegraphics[width=5.5cm]{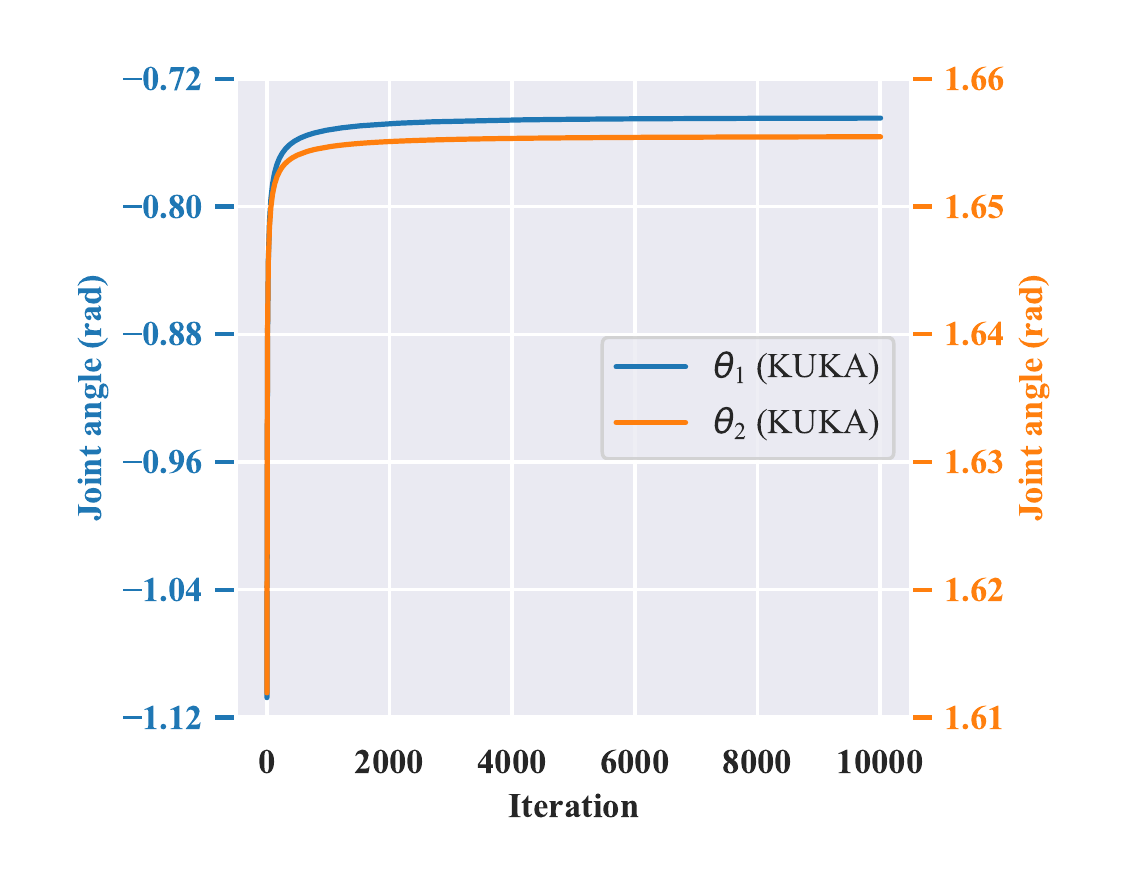}}
	\hspace{-0.3cm}
	\subcaptionbox{\label{comp_3}}[5.5cm][c]{\includegraphics[width=5.5cm]{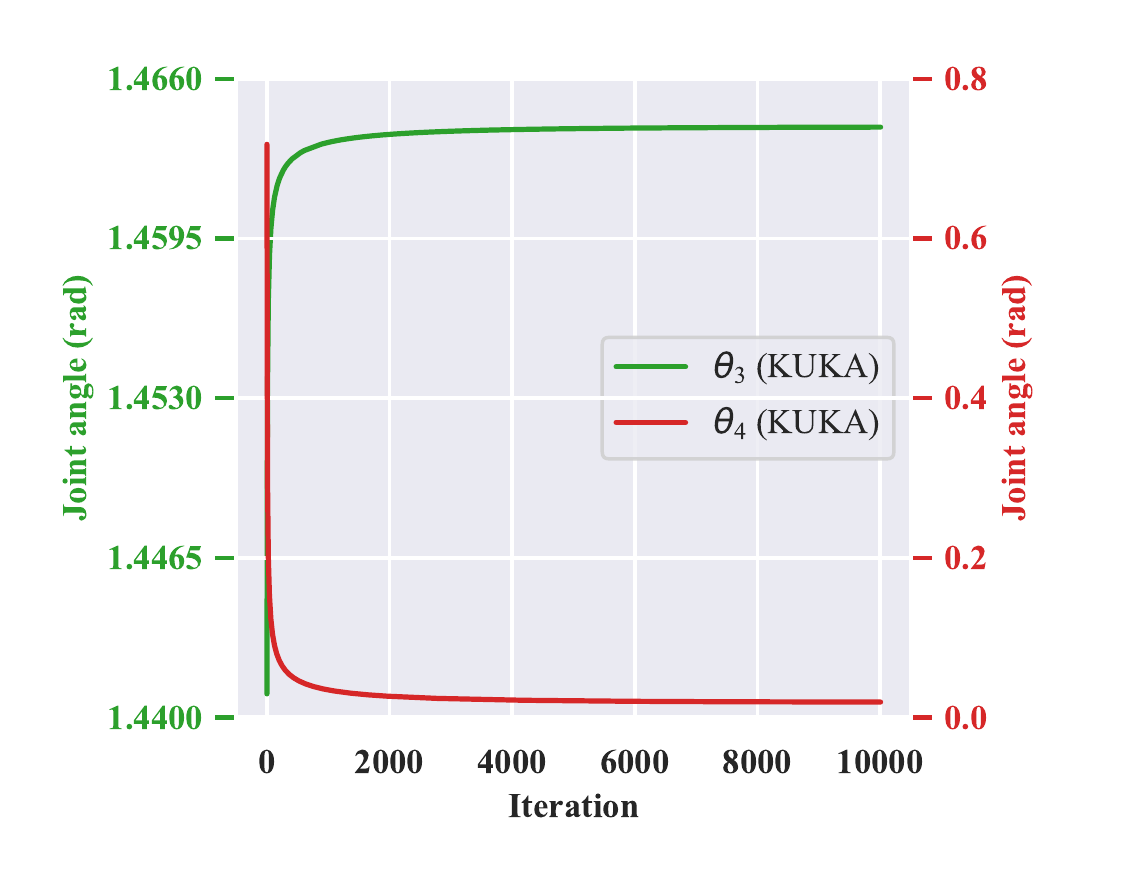}}
	\vskip -5pt
	\subcaptionbox{\label{comp_4}}[5.5cm][c]{\includegraphics[width=5.5cm]{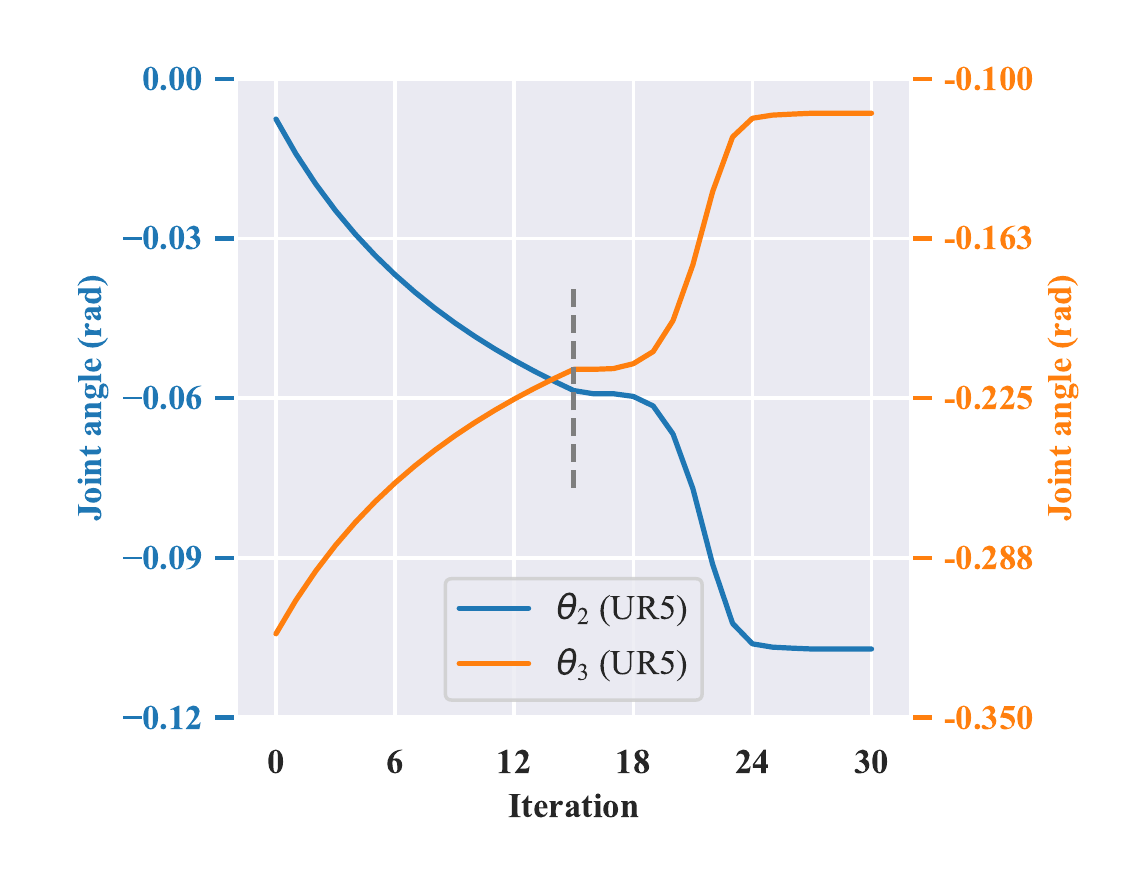}}
	\hspace{-0.3cm}
	\subcaptionbox{\label{comp_5}}[5.5cm][c]{\includegraphics[width=5.5cm]{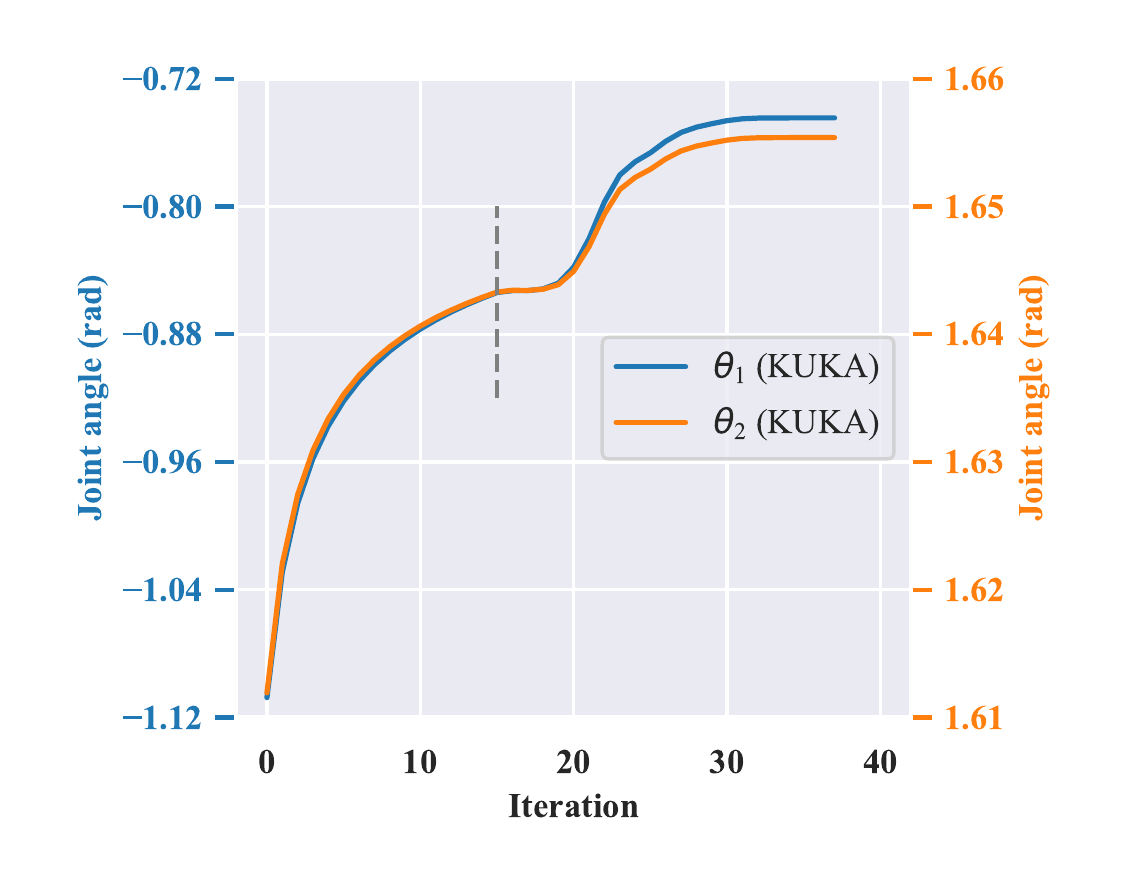}}
	\hspace{-0.3cm}
	\subcaptionbox{\label{comp_6}}[5.5cm][c]{\includegraphics[width=5.5cm]{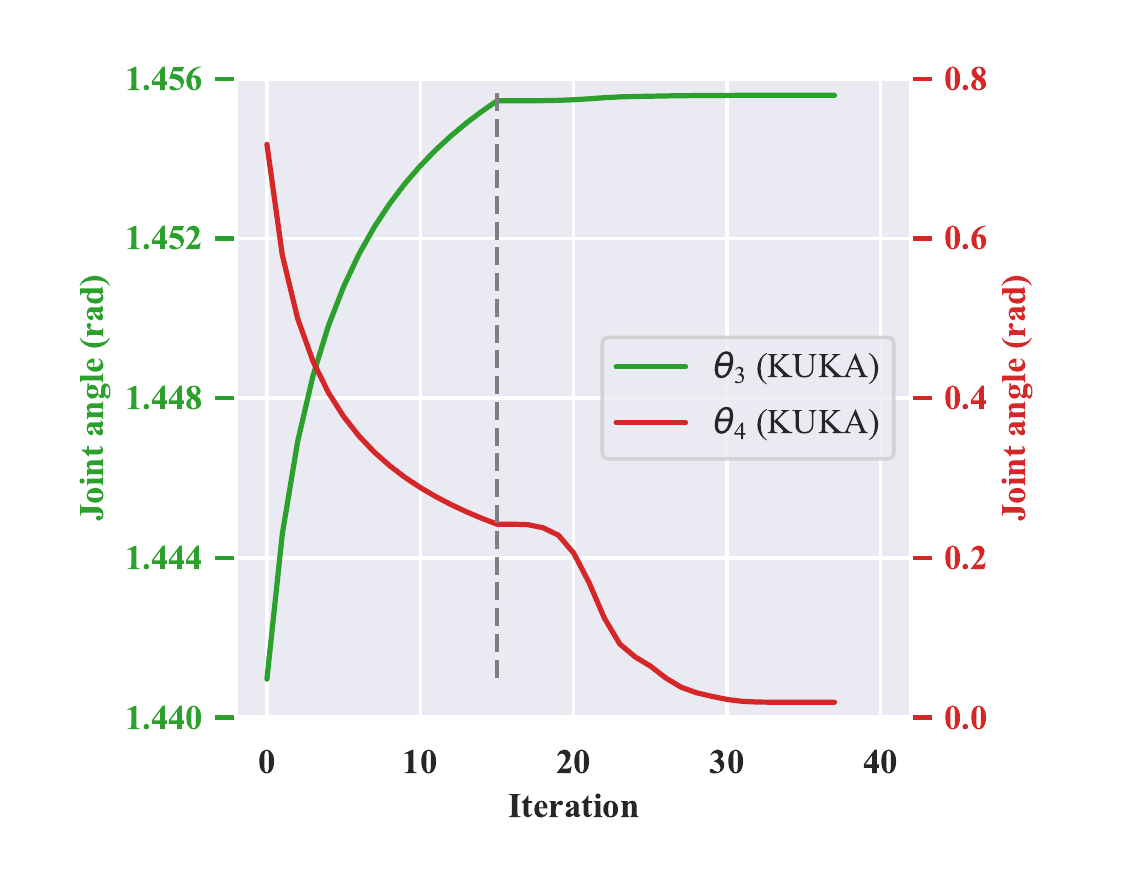}}
	%	\subcaptionbox{\label{comp_7}}[4.4cm][c]{\includegraphics[width=4.4cm]{ur5_opt_dist.pdf}}
	%	\hspace{-0.3cm}
	%	\subcaptionbox{\label{comp_8}}[4.4cm][c]{\includegraphics[width=4.4cm]{kuka_opt_dist.pdf}}
	\caption{Convergence experiment results comparing the combined algorithm with FABRIK, where the variations in joint angles of the UR5 and KUKA manipulators generated by FABRIK and the combined algorithm are presented in the first and second rows, respectively.}
	\label{UR5_convergence_comp}
\end{figure*}
\subsection{Comparison of Convergence}
%%%%%%%%%%%%%%%%%%%%%%%%%%%%%%%%%
\begin{table*}[h!]
	\caption{QUANTITATIVE COMPARISON RESULTS OF THE COMBINED ALGORITHM AND FABRIK WITH VARIOUS LIMITS ON THE UR5 AND KUKA MANIPULATORS}
	\label{success_time}
	\renewcommand\arraystretch{1.2}  % 修改行距
	\begin{center}
%		\resizebox{\textwidth}{!}
%		{
			\begin{tabular}{C{1.3cm} C{0.9cm} C{0.9cm} C{0.9cm} C{0.9cm} C{0.9cm} C{0.9cm} C{0.9cm} C{0.8cm} C{0.9cm} C{0.9cm} C{0.9cm} C{0.9cm}}
				\toprule
				Algorithms & \multicolumn{2}{c}{\makecell[c]{Combined \\ ($ n_l = 5 $)}} & \multicolumn{2}{c}{\makecell[c]{Combined \\ ($ n_l = 15 $)}} & \multicolumn{2}{c}{\makecell[c]{Combined \\ ($ n_l = 50 $)}} & \multicolumn{2}{c}{\makecell[c]{FABRIK \\ ($ n_{max} = 100 $)}}& \multicolumn{2}{c}{\makecell[c]{FABRIK \\ ($ n_{max} = 500 $)}} & \multicolumn{2}{c}{\makecell[c]{FABRIK \\ ($ n_{max} = 900 $)}} \\ [0.5ex]
				\cline{2-13}
				\rule{0pt}{2.6ex}    
				{} & Avg. Time (ms)  & Succ. Rate (\%) & Avg. Time (ms)  & Succ. Rate (\%) & Avg. Time (ms)  & Succ. Rate (\%) & Avg. Time (ms) & Succ. Rate (\%) & Avg. Time (ms) & Succ. Rate (\%) & Avg. Time (ms) & Succ. Rate (\%) \\ 
				\midrule
				UR5 & 0.442 & 99.99 & 0.668 & 99.99 & 1.590 & 99.99 & 2.084 & 95.31 & 3.704 & 98.88 & 4.659 & 99.23 \\
				KUKA & 0.278 & 99.64 & 0.359 & 99.83 & 0.506 & 99.89 & 0.630 & 85.63 & 1.297 & 93.97 & 1.605 & 95.73  \\
				\bottomrule 
			\end{tabular}
%		}
	\end{center}
\end{table*}
%%%%%%%%%%%%%%%%%%%%%%%%%%%
In order to evaluate the convergence of FABRIK and the combined algorithm under the $\varepsilon_{tol} = 10^{-6}$ constraint, two IK requirements
\begin{equation}
	\label{given_targets}
	\begin{array}{l}
	T_{des}^{6} = \left[ {\begin{array}{*{20}{c}}
			{{\rm{ - 0}}.{\rm{770}}}&{{\rm{0}}.{\rm{618}}}&{{\rm{0}}.{\rm{156}}}&{{\rm{ - 0}}.{\rm{296}}}\\
			{{\rm{ - 0}}.{\rm{638}}}&{{\rm{ - 0}}.{\rm{740}}}&{{\rm{ - 0}}.{\rm{214}}}&{{\rm{ - 0}}.{\rm{869}}}\\
			{{\rm{ - 0}}.{\rm{017}}}&{{\rm{ - 0}}.{\rm{265}}}&{{\rm{0}}.{\rm{964}}}&{{\rm{0}}.{\rm{288}}}\\
			0&0&1&1
	\end{array}} \right] \textrm{and} \nonumber
	\end{array}
\end{equation}
%%%%%%%%%%%%%%%%%%%%%%%%%%%
\begin{figure}[t!]
	\centering
	\includegraphics[width=8.7cm]{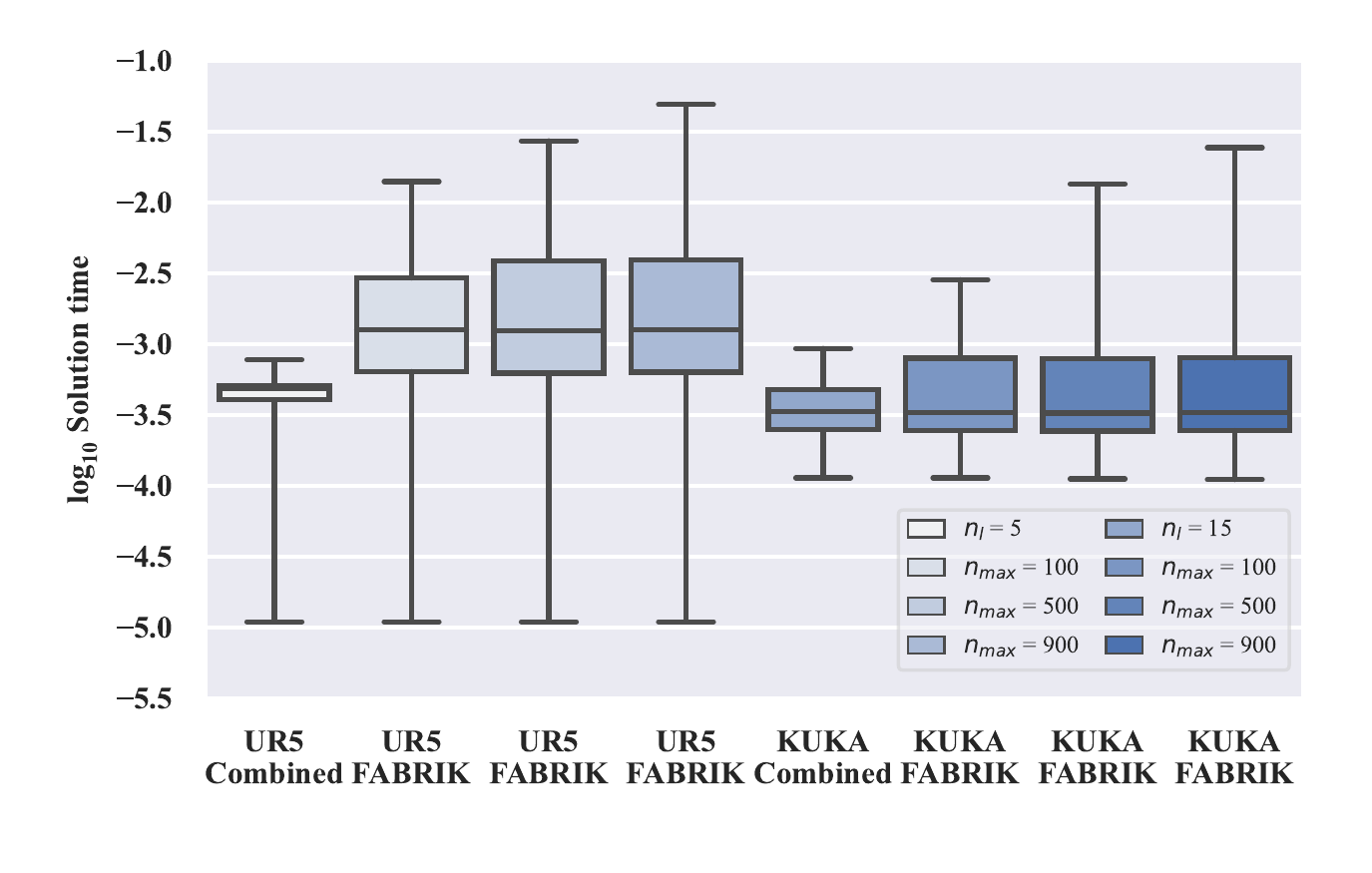}
	%	\subcaptionbox{\label{plot_2}}[7cm][c]{\includegraphics[width=7cm]{barplot.pdf}}
	\caption{Boxplot summarizing the logarithm of the solution time of the combined algorithm and FABRIK with various $ n_l $ and $ n_{max} $ in the random test.}
	\label{plot_bar}
	\vskip -13pt
\end{figure}
%%%%%%%%%%%%%%%%%%%%%%%%%%%%
\begin{equation}
	\label{given_targets}
	\begin{array}{l}
		T_{des}^{7} = \left[ {\begin{array}{*{20}{c}}
				{{\rm{0}}{\rm{.537}}}&{{\rm{0}}{\rm{.838}}}&{{\rm{0}}{\rm{.094}}}&{{\rm{0}}{\rm{.618}}}\\
				{{\rm{ - 0}}{\rm{.654}}}&{{\rm{0}}{\rm{.344}}}&{{\rm{0}}{\rm{.674}}}&{{\rm{ - 0}}{\rm{.463}}}\\
				{{\rm{0}}{\rm{.532}}}&{{\rm{ - 0}}{\rm{.423}}}&{{\rm{0}}{\rm{.733}}}&{{\rm{0}}{\rm{.382}}}\\
				0&0&0&1 \nonumber
		\end{array}} \right]
	\end{array}
\end{equation}
are specified for the UR5 and KUKA manipulators, with $ n_l = 15 $ in this subsection. For the UR5 manipulator, after determining $ \bm{P}_t $ using Eq.~\eqref{ur5_target}, the iteration process that happens in $ \bm{n_p} $ can be recorded as the variations of $\theta_2$ and $\theta_3$. As shown in the comparison results in Figs.~\ref{comp_1} and \ref{comp_4}, FABRIK spans 546 iterations to converge within joint limits. In contrast, the combined algorithm only takes 30 iterations to converge, including 15 optimization steps. The gray dotted lines in Figs.~\ref{comp_4}-\ref{comp_6} denote the beginning of the optimization phase of the combined algorithm. Analogously, for the KUKA manipulator, $ \theta_1 $,  $ \theta_2 $, and $ \theta_4 $, which are deduced by the combined algorithm, can converge to the same limits as those generated by FABRIK. Whether using the combined algorithm or FABRIK, the limit to which $ \theta_3 $ converges has no effect on the position of $ \bm{P}_3 $. Compared to the 10010 iterations consumed by FABRIK, the experiment results in Figs.~\ref{comp_5}-\ref{comp_6} further demonstrate that the combined algorithm can achieve faster convergence when applied to the KUKA manipulator. The convergence behaviors shown in Figs.~\ref{comp_4}-\ref{comp_6} also verify the feasibility of the combined algorithm's switch index. 

Finally, the desired joint angle vectors, which are derived by the combined algorithm for the UR5 and KUKA manipulators, are $ \bm{\Theta}_{des}^{6} = $ $ \big[$1.103, -0.107, -0.114, -1.226, 1.333, -1.995$ \big]^T$rad and $ \bm{\Theta}_{des}^{7} = $ $\big[$-0.745, 1.655, -1.686, -0.019, 1.003, -2.025, -0.505$\big]^T$rad, respectively. Additionally, the convergence limits of $\theta_3$ and $\theta_4$ in Fig.~\ref{comp_6} differ from corresponding values in $ \bm{\Theta}_{des}^{7} $ but with the same $ \left\{ \widehat {{\bm{l}_{2d}}} ,\widehat {{\bm{l}_{3d}}}\right\} $, demonstrating the necessity to recalculate all joint angles after optimization.
\begin{figure*}[t]
	\centering
	\captionsetup[subfigure] {skip=5pt,slc=off,margin={65pt, 0pt},labelfont=normalfont}
	\subcaptionbox{\label{tracking_1}}[4.5118cm][c]{\includegraphics[width=4.5cm]{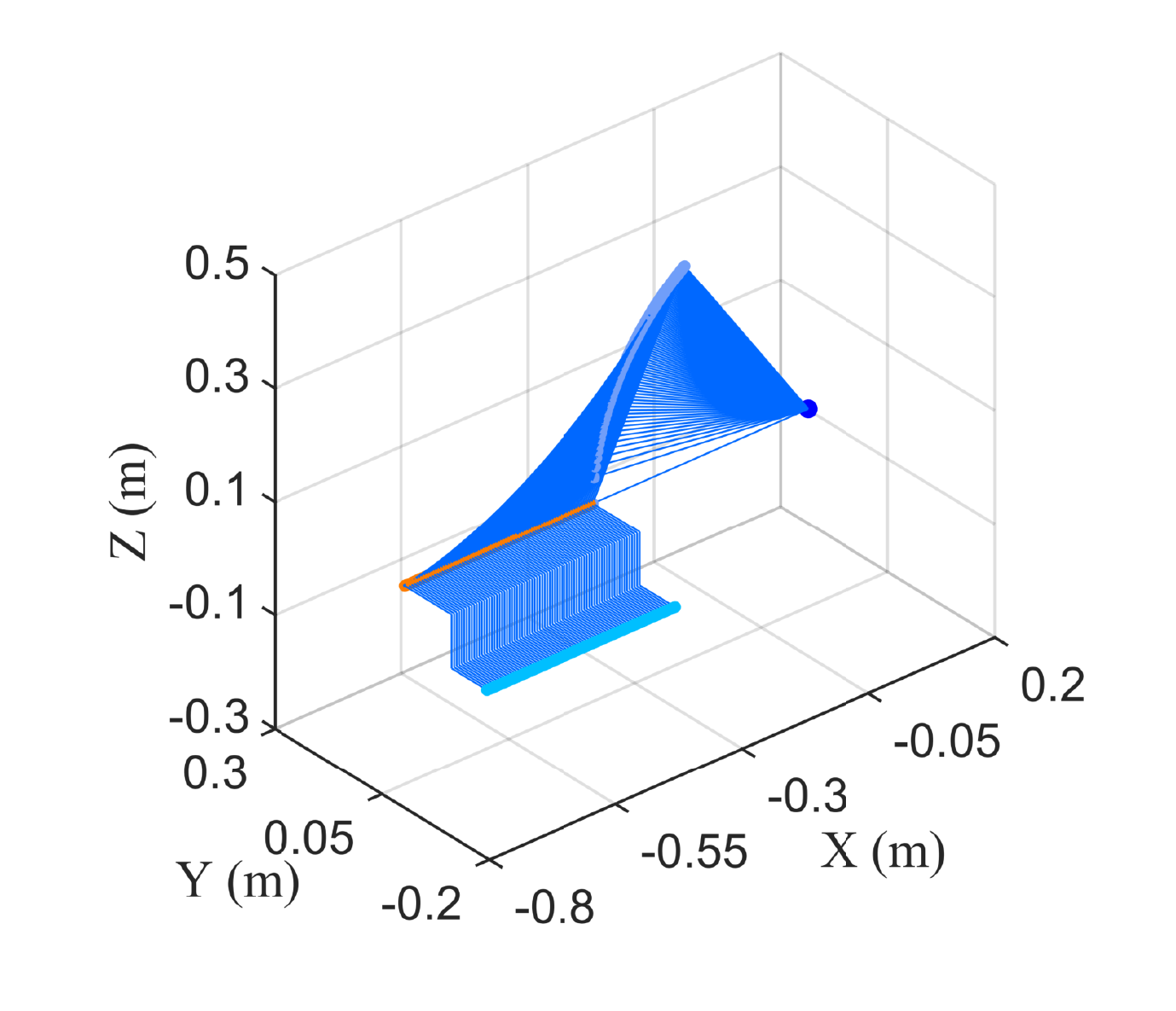}} \subcaptionbox{\label{tracking_2}}[4.5118cm][c]{\includegraphics[width=4.5cm]{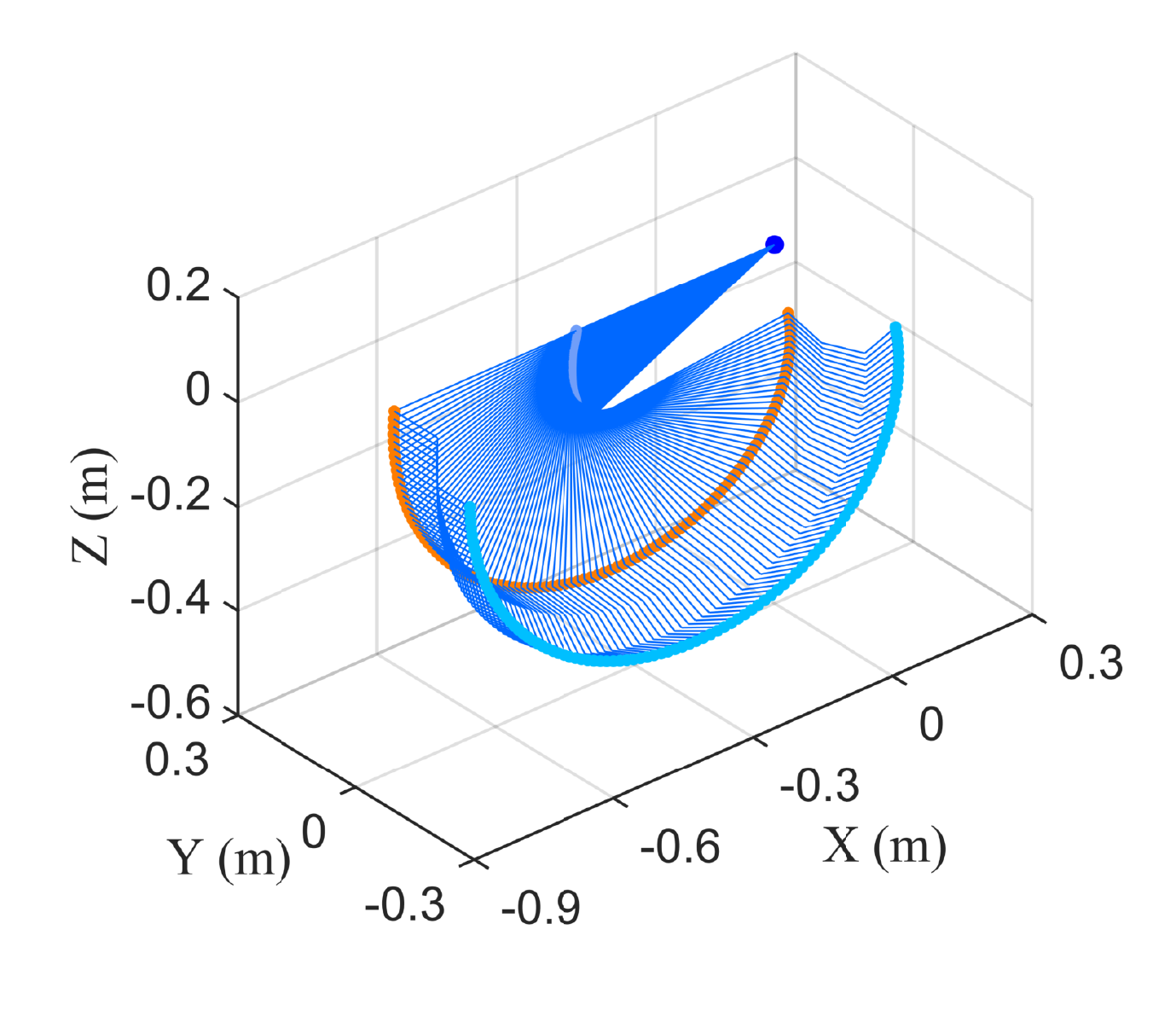}}
	\captionsetup[subfigure] {skip=5pt,slc=off,margin={60pt, 0pt},labelfont=normalfont} 
	\subcaptionbox{\label{tracking_3}}[4cm][c]{\includegraphics[width=4.5cm]{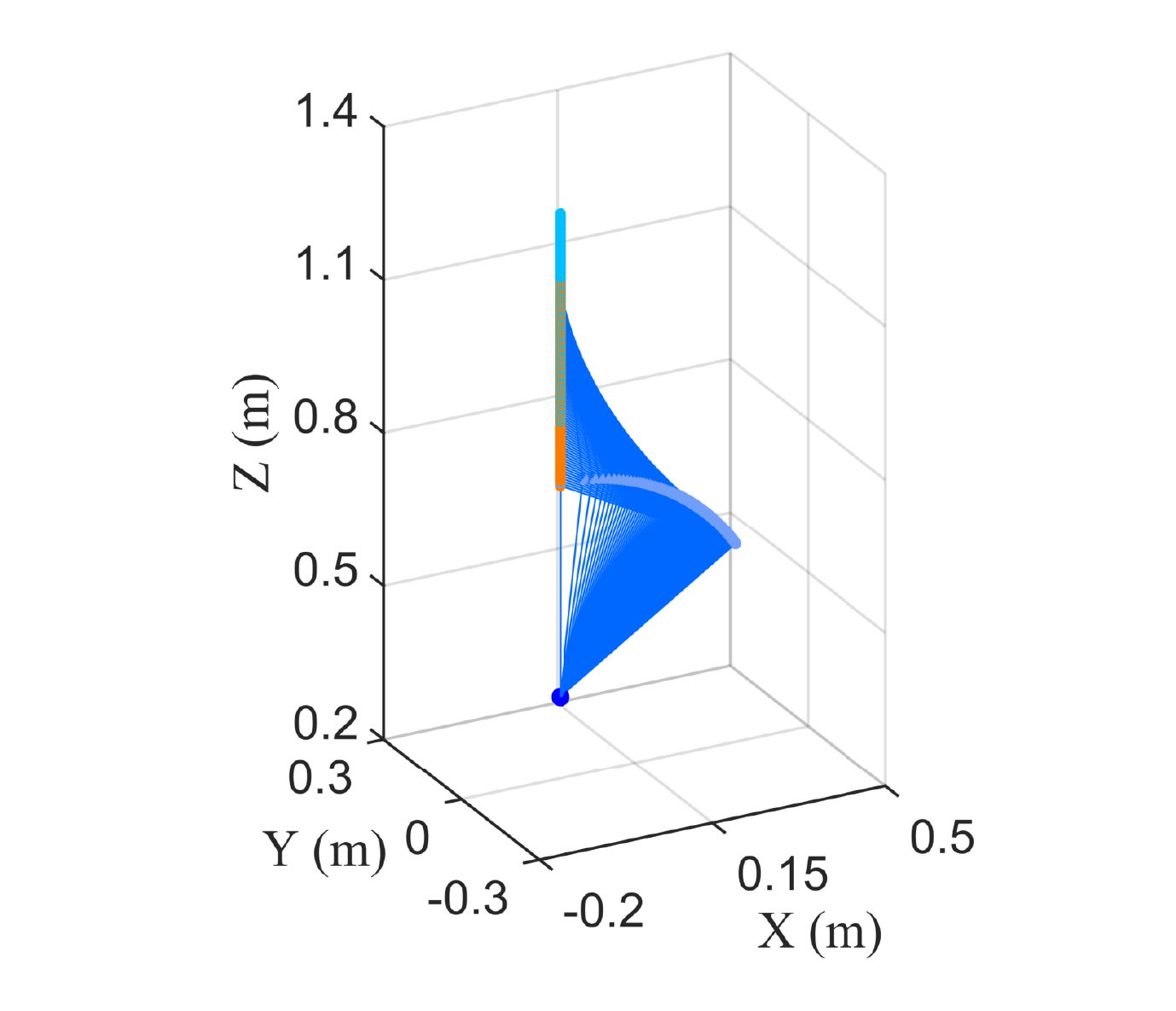}} \subcaptionbox{\label{tracking_4}}[4cm][c]{\includegraphics[width=4.5cm]{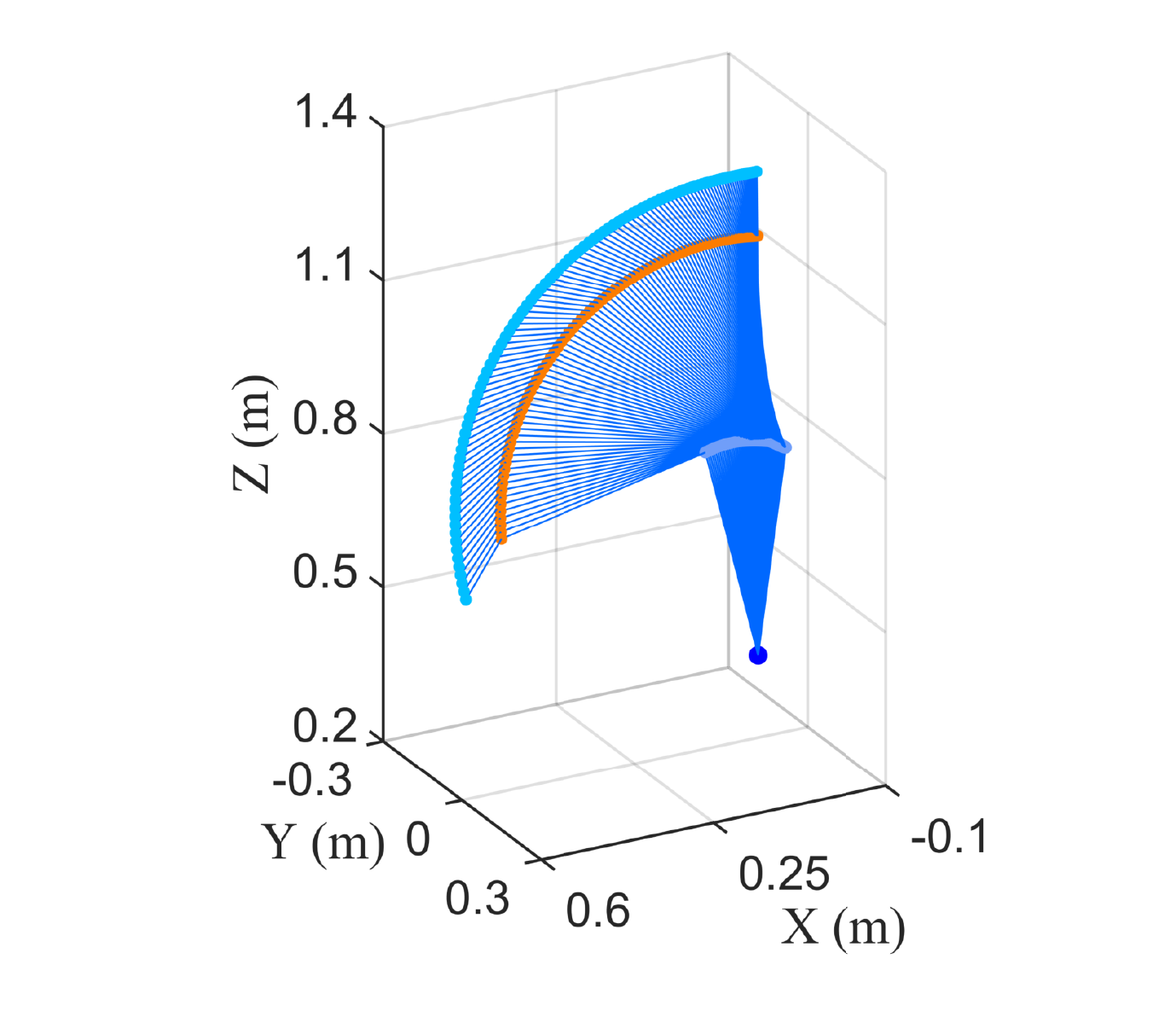}}
	\caption{The first and second phase motion trajectories of the (\subref{tracking_1}) (\subref{tracking_2}) UR5 and (\subref{tracking_3}) (\subref{tracking_4}) KUKA manipulators, where \tikzcircle[blue, fill=blue]{3pt} $ \bm{P}_1 $ and the paths of \tikzcircle[elbow1_color,fill=elbow1_color]{3pt} $ \bm{P}_2 $, \tikzcircle[elbow2_color, fill=elbow2_color]{3pt} $ \bm{P}_3 $, and \tikzcircle[ee_color, fill=ee_color]{3pt} EE are indicated. In (\subref{tracking_1}) and (\subref{tracking_3}), $ \bm{P}_3 $ tracks the path points on $ \bm{v}_{init} $ until the kinematic chains reach the zero positions.} 
	\label{path_tracking}
\end{figure*}

\subsection{Quantitative Evaluation}
The average solution time and success rate of the combined algorithm and FABRIK are quantitatively evaluated on the UR5 and KUKA manipulators with 10,000 random IK queries and the  $\varepsilon_{tol} = 10^{-6}$ error constraint, where the combined algorithm and FABRIK are performed with different $ n_{l} $ and $ n_{max} $. Tab.~\ref{success_time} summarizes the random test results, in which the combined algorithm outperforms FABRIK in terms of success rate and average solution time only with $ n_l = 5 $, especially when applied to the KUKA manipulator. Although both the kinematic chains of the UR5 and KUKA manipulators involved in iterations only have two links, the 2-D iterations could be performed up to four times due to the multiple iteration targets deduced by Eqs.~\eqref{UR5-2} and \eqref{UR5-3}. For the 2-D scenario, the combined algorithm has exhibited the best performance when $ n_l = 5 $. In contrast, $ n_l = 15 $ is the best option for the combined algorithm after weighting the average solution time and success rate in the 3-D case. The success rate of FABRIK increases with $ n_{max} $ but with significantly higher time costs, indicating that the time-consuming iterations of FABRIK occur. Once $ n_{max} $ exceeds a certain value, the increase of $ n_{max} $ on the success rate of FABRIK is limited. Thus, FABRIK may take an unpredictable long time to complete the inefficient iterations to improve the success rate even further. The boxplot in Fig.~\ref{plot_bar} displays the interquartile range, minimum and maximum logarithms of the solution time of the combined algorithm and FABRIK, where the switch index of the combined algorithm is 5 and 15 in the 2-D and 3-D cases, respectively. The maximum solution time of FABRIK also increases with $ n_{max} $. In contrast to FABRIK, the combined algorithm generates a narrower performance range and a solution time distribution that concentrates in the region with less computation cost. Note that, due to the fast convergence characteristics, FABRIK can complete the calculation in some cases within $ n_l $, as illustrated by the minimum values of the solution time for various tests in Fig.~\ref{plot_bar}. Additionally, the wider performance range and average solution time of FABRIK also demonstrate the necessity of optimization.

\subsection{Path Tracking}
\begin{figure*}[th]
	\centering
	\captionsetup[subfigure] {skip=5pt,slc=off,margin={68pt, 0pt},labelfont=normalfont}
	\subcaptionbox{\label{angle_error_1}}[4.5cm][c]{\includegraphics[width=4.5cm]{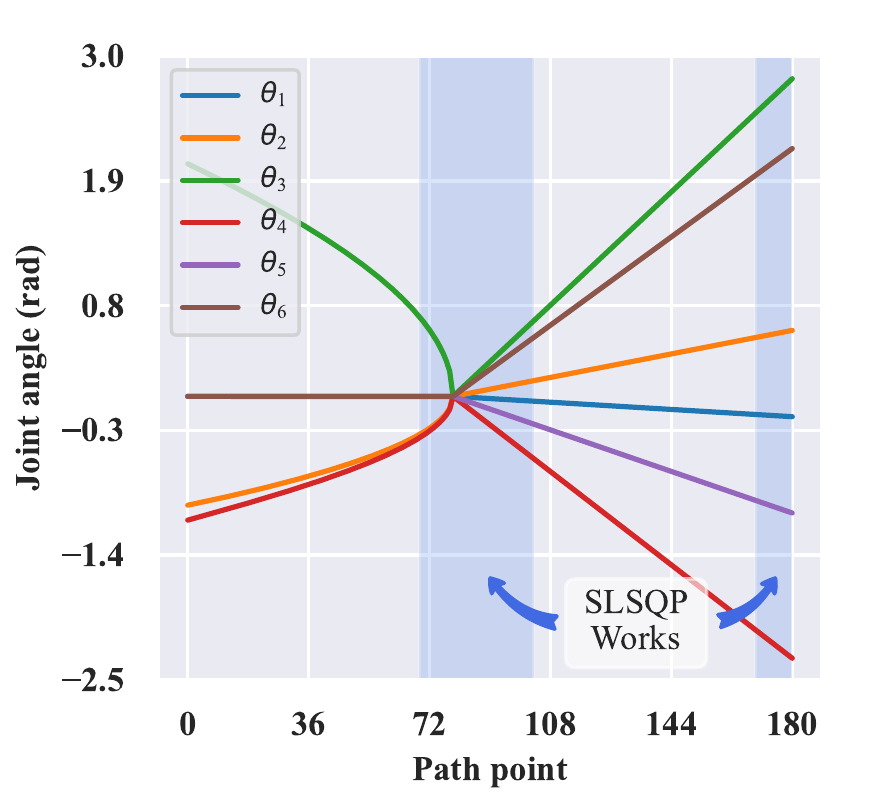}}
	\hspace{-0.3cm}
	\subcaptionbox{\label{angle_error_2}}[4.5cm][c]{\includegraphics[width=4.5cm]{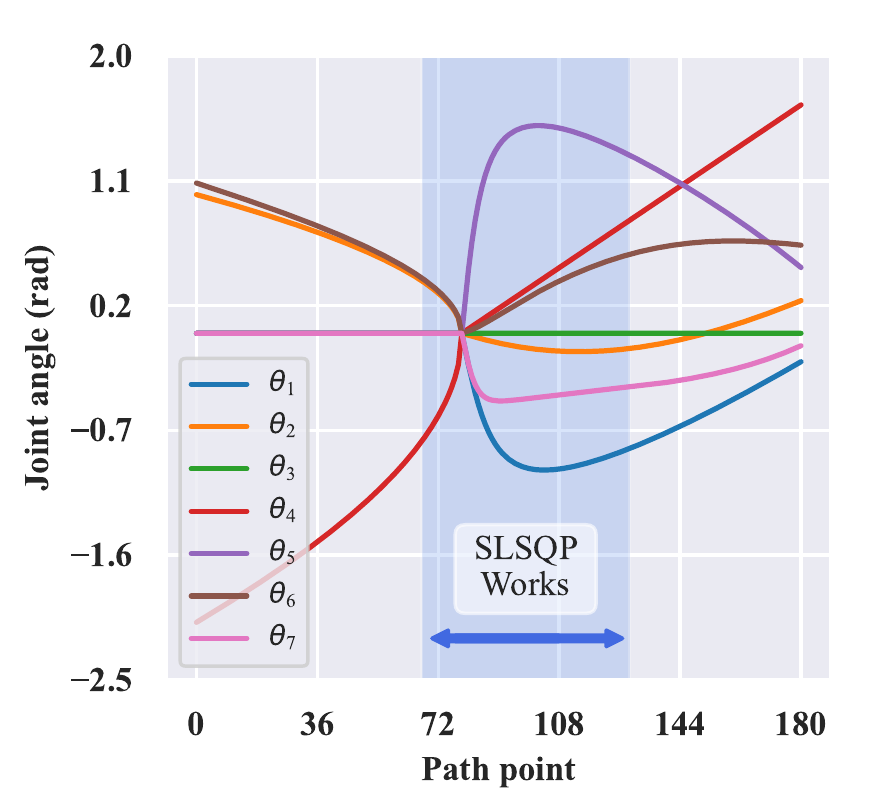}}
	\hspace{-0.2cm}
	\subcaptionbox{\label{angle_error_3}}[4.5cm][c]{\includegraphics[width=4.5cm]{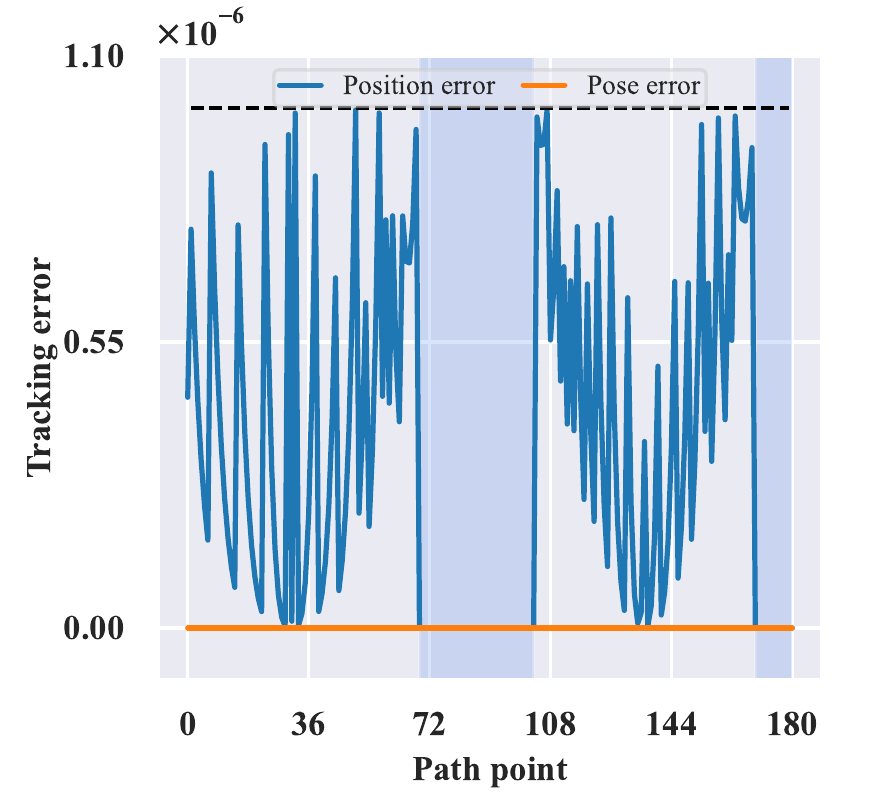}}
	\hspace{-0.6cm} \subcaptionbox{\label{angle_error_4}}[4.5cm][c]{\includegraphics[width=4.5cm]{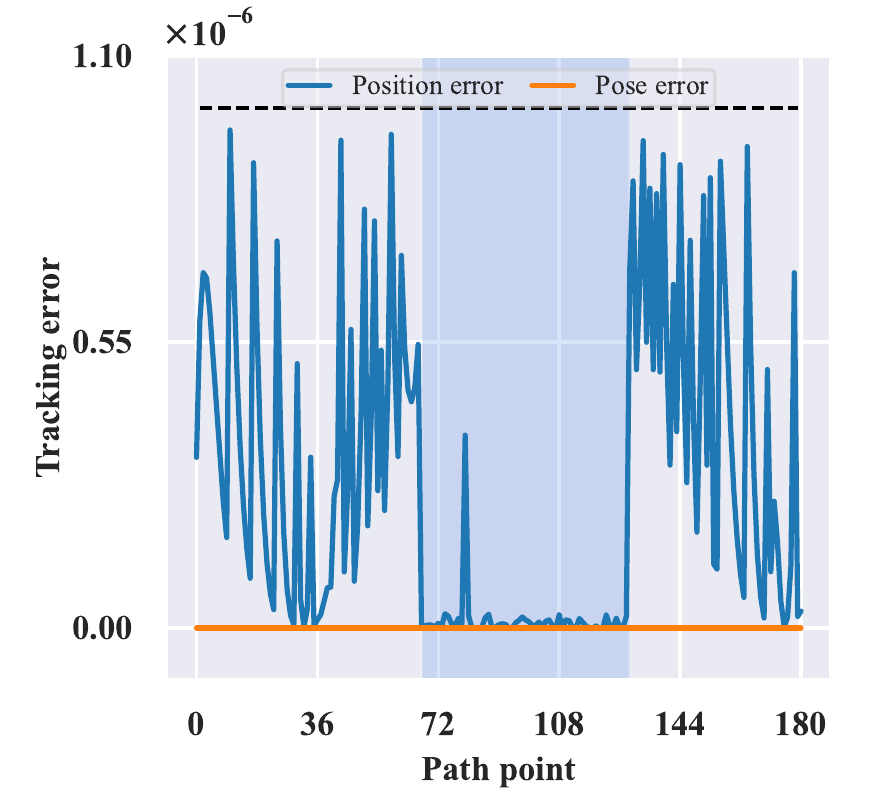}}
	\caption{Joint angle and Cartesian error variations of the (\subref{angle_error_1}) (\subref{angle_error_3}) UR5 and (\subref{angle_error_2}) (\subref{angle_error_4}) KUKA manipulators, where the activation of SLSQP is emphasized. The dashed lines in (\subref{angle_error_3}) and (\subref{angle_error_4}) indicate the given Cartesian error constraint. } 
	\label{path_tracking_angle_error}
	\vskip -2pt
\end{figure*}
The path tracking task in this subsection is divided into two phases, in which $ \bm{P}_3 $ should continuously track the path points that lie on the same line as $ \bm{v}_{init} $ in the first phase, and EE should follow a random path in the second phase. The random path is formed by starting at the zero position and randomly selecting a configuration as the end point, then interpolating in the configuration space and obtaining the path points using forward kinematics. The manipulators should trace 80 and 100 path points in two phases with the $\varepsilon_{tol} = 10^{-6}$ error constraint.

The UR5 and KUKA manipulators begin with the initial configurations of $ \bm{\Theta}_{init}^6$ = $\big[$0, -0.959, 2.05, -1.091, 0, 0$\big]^T$rad and $ \bm{\Theta}_{init}^7$ = $\big[$0, 1.000, 0, -2.084, 0, 1.084, 0$\big]^T$rad, respectively. Using forward kinematics, two randomly selected configurations, $ \bm{\Theta}_{end}^6$ = $\big[$-0.179, 0.581, 2.8, -2.308, -1.028, 2.185$\big]^T$rad and $ \bm{\Theta}_{end}^7$ = $\big[$1.953, -0.711, -1.608, 1.648, -0.888, 0.782, 0.893$\big]^T$rad, are utilized to deduce the end of two paths. The two-phase tracking processes are shown in Fig.~\ref{path_tracking}. When using the combined algorithm, both manipulators can complete the tracking tasks with continuous configurations, and $ \bm{P}_3 $ tracks the targets on $ \bm{v}_{init} $ as anticipated in the first phase. Figs.~\ref{angle_error_1} and \ref{angle_error_2} depict the joint angle profiles of the UR5 and KUKA manipulators deduced by the combined algorithm, in which the activation of SLSQP during path tracking is displayed. Notably, although bending the kinematic chain before iteration as described in \cite{aristidou2016extending} to avoid endless loops, SLSQP is still activated at the end of the first phase and the beginning of the second phase. Meanwhile, the UR5 manipulator must bend significantly at the end of the second phase. These two phenomena further demonstrate that FABRIK tends to trap in inefficient iterations to slowly update joint positions when the kinematic chain must bend slightly or dramatically to achieve the target under the high error constraint. For the tracking tasks of the UR5 and KUKA manipulators, the average solution time of the combined algorithm is 0.529 ms and 0.395 ms, respectively, which indicates the effectiveness of the combination of FABRIK and SLSQP and can provide real-time motions. The tracking errors generated by the combined algorithm are shown in Figs.~\ref{angle_error_3} and \ref{angle_error_4}, from which it can be observed that the combined algorithm does not induce the pose error of EE and the generated position errors are less than $ \varepsilon_{tol} $. Especially when SLSQP is activated, the position error can be further decreased to almost zero.

%%%%%%%%%%%%%%%%%%%%%%%%%%%%%%%%%%%%%%%%%%%%%%%%%%%%%%%%%%%%%%%%%%%%%%%%%%%%%%%%
\section{Conclusion and future work}
\label{sec6}
In this article, a novel combined algorithm is presented for applying FABRIK to manipulators and optimizing the unstable convergence property of FABRIK. The combination of FABRIK and the SQP algorithm substantially prevents FABRIK from getting stuck in inefficient iterations. The convergence comparison experiment showed that the combined algorithm converges faster than FABRIK and its switch condition is feasible. The quantitative experiment also demonstrated that the combined algorithm outperforms FABRIK in terms of solution time and success rate when applied to the UR5 and KUKA manipulators. Using the combined algorithm, both the UR5 and KUKA manipulators can complete the tracking tasks with continuous configurations, zero pose error and permitted position error of EE. The convergence analysis and tracking results concluded that FABRIK will be prone to inefficient iterations when the kinematic chain needs to bend slightly or significantly to reach the target. Overall, the combined algorithm fully exploits the advantages of FABRIK and the SQP algorithm and achieves better computational performance under the high error constraint.

Future work will focus on extending the combined algorithm to more manipulators with different structures. More work is required to apply FABRIK to the manipulators with multiple joint offsets and complex joint limits to provide excellent performance under high error constraints. Some additional goals, such as collision avoidance in 3-D space, can be realized by modifying the iteration process of FABRIK to take advantage of the redundancy of some manipulators. 
%%%%%%%%%%%%%%%%%%%%%%%%%%%%%%%%%%%%%%%%%
%\bibliography{ref_2022_5_19}

\end{document}